\documentclass[10pt,twocolumn,letterpaper]{article}

\usepackage{iccv}              %

\usepackage[accsupp]{axessibility}  %

\definecolor{iccvblue}{rgb}{0.21,0.49,0.74}
\usepackage[pagebackref,breaklinks,colorlinks,allcolors=iccvblue]{hyperref}

\usepackage{times}
\usepackage{epsfig}
\usepackage{graphicx}
\usepackage{amsmath}
\usepackage{amssymb}
\usepackage{amsfonts}       %

\usepackage{url}            %
\usepackage{booktabs}       %
\usepackage{nicefrac}       %
\usepackage{microtype}      %
\usepackage{xcolor}         %
\usepackage{mathabx}        %
\usepackage{wrapfig}        %
\usepackage{adjustbox}
\usepackage{multirow}

\usepackage{algorithm}
\usepackage{algpseudocode}

\usepackage{comment}
\usepackage{nth}
\usepackage{enumitem}
\usepackage{tikz}
\usepackage{pgfplots}

\def\paperID{14982}
\def\confName{ICCV}
\def\confYear{2025}

\newcommand{\ourmetric}{GFLOPs}
\newcommand{\Ourmetric}{GFLOPs}
\newcommand{\Ourmethod}{our method}
\newcommand{\topic}[1]{\noindent\textbf{#1}}
\newcommand{\figspacebottom}{-1.0em}

\def\CircleArrowright{\ensuremath{%
  \rotatebox[origin=c]{310}{$\circlearrowright$}}}
\newcommand{\vlncebert}{VLN-CE$\protect\CircleArrowright$BERT }

\begin{document}

\title{Harnessing Input-Adaptive Inference for Efficient VLN}

\author{%
    Dongwoo Kang, Akhil Perincherry, Zachary Coalson, Aiden Gabriel, Stefan Lee, Sanghyun Hong \\
    \textit{Oregon State University} \\
    {\tt \small \{kangdo, perincha, coalsonz, gabrieai, leestef, sanghyun.hong\}@oregonstate.edu} %
}

\maketitle

\begin{abstract}
    An emerging paradigm in vision-and-language navigation (VLN)
    is the use of history-aware multi-modal transformer models. 
    Given a language instruction, 
    these models process observation and navigation history
    to predict the most appropriate action for an agent.
    While they have significantly improved performance,
    the scale of these models can be a bottleneck in practical settings
    with limited computational resources.
    In this work, we propose a novel input-adaptive navigation method
    to enhance VLN model efficiency.
    We first show that existing input-adaptive mechanisms 
    fail to reduce computations without substantial performance degradation.
    To address this, we introduce three adaptive algorithms,
    each deployed at a different level:
    (1) To improve spatial efficiency,
    we selectively process panoramic views at each observation of an agent.
    (2) To improve intra-model efficiency,
    we propose importance-based adaptive thresholding for the early-exit methods.
    (3) To improve temporal efficiency,
    we implement a caching mechanism that prevents reprocessing of views 
    previously seen by the agent.
    In evaluations on seven VLN benchmarks,
    we demonstrate over a 2$\times$ reduction in computation
    across three off-the-shelf agents in both standard and continuous environments. 
    Our code is publicly available at \url{https://github.com/secure-ai-systems-group/adaptive-vision-and-language-navigation}.
\end{abstract}

\section{Introduction}
\label{sec:intro}

Progress in vision-and-language navigation (VLN) has been enabled by larger models
trained on increasingly large datasets%
~\cite{hao2020towards, HAMT, moudgil2021soat, guhur2021airbert, vlnbert}.
These models can process and interpret complex data,
enabling them to understand and act upon natural language instructions 
within visual environments.
Despite the success, there is a growing concern 
about their computational demands.
The need for substantial computational power 
poses a notable challenge for deployment in 
resource-constrained settings, such as robots,
where low-power consumption becomes increasingly critical.

A potential solution to addressing these computational demands 
is \emph{input-adaptive inference}.
The main idea is to reduce \emph{overthinking}~\cite{kaya2019shallow}:
as shallow networks are sufficient for the majority of samples 
to make decisions, e.g., class predictions,
input-adaptive methods~\cite{msdnet, liu2020fastbert, xin2020deebert, MuE}                        %
stop forwarding preemptively during inference and return intermediate outputs
when the internal decisions of a model converge.
During inference, they demonstrate
up to 50\% computational savings
while preserving model performance.

In this work, we study the overthinking problem in a new domain%
---VLN---and
propose a novel input-adaptive method to address it.
Unlike prior studies on overthinking,
which focus on tasks where inputs are processed independently (e.g., classification),
VLN involves sequential decision-making, introducing unique problems 
driven by \emph{spatio-temporal dependencies} in the inputs.
Moreover, these models can be deployed for real-world navigation;
thus, it is important to assess whether they are robust to common visual corruptions.

\begin{figure*}[t] 
\vspace{-0.2em}     %
\centering
\includegraphics[width=\linewidth]{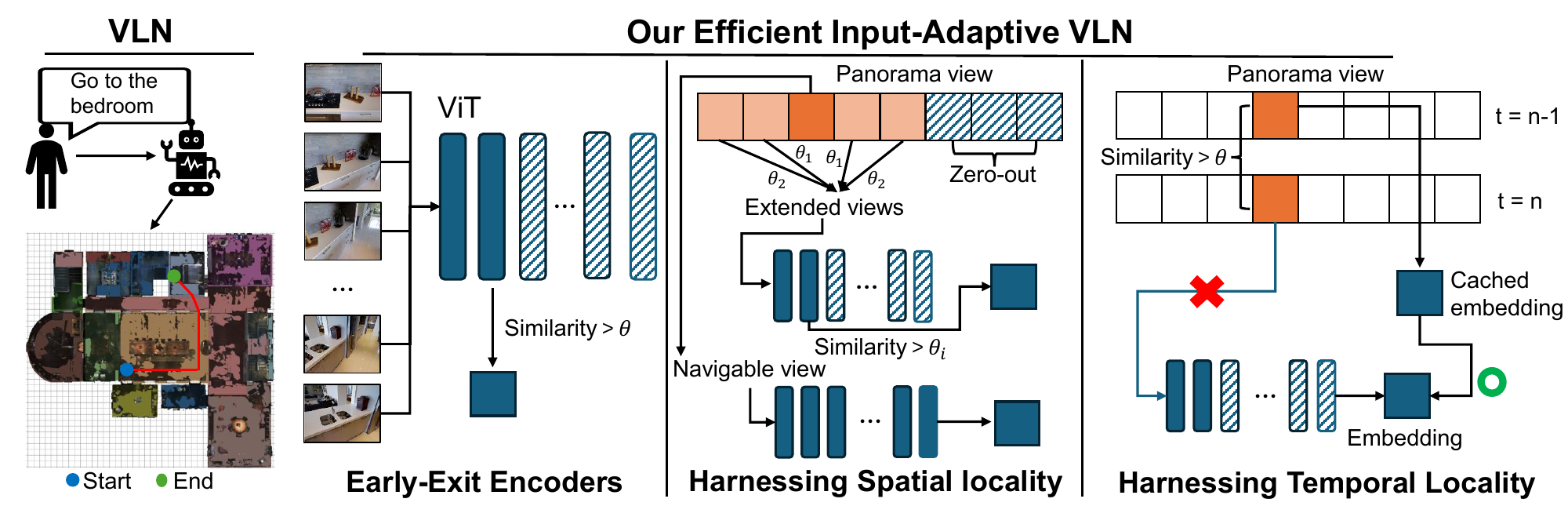}
\vspace{-2em}
\caption{%
    \textbf{Our input-adaptive, efficient navigation method.} We show on the left an agent navigating a visual environment upon a natural language instruction. On the right, we provide a high-level overview of the three input-adaptive mechanisms we propose at different levels. The shaded rectangles (embeddings) and squares (views) correspond to components that our method skips or zeroes out to improve efficiency.}
\label{fig:our-method-overview}
\vspace{\figspacebottom}
\end{figure*}

\topic{Contributions.}
We \emph{first} characterize the overthinking problem in VLN
by analyzing its computational bottlenecks.
In our evaluation with two standard VLN agents (HAMT~\cite{HAMT} and DUET~\cite{chen2022think}) 
and one continuous VLN agent (\vlncebert~\cite{krantz2022sim}),
we find that $\sim$99.5\% of computations are spent in visual encoders.
We also show that addressing overthinking 
within these visual encoders is ineffective in
providing computational savings.
Even with our best effort to apply 
the existing input-adaptive inference method, MuE~\cite{MuE},
we demonstrate that this approach results in inaccurate navigation decisions.
This increases both the time it takes for an agent 
to reach the target location and the overall computations 
while lowering the navigation success.

\emph{Second}, to address this issue and achieve computational efficiency,
we propose a novel input-adaptive navigation method 
(shown in Figure~\ref{fig:our-method-overview}).
We not only minimize overthinking within visual encoders,
as in prior approaches, but also reduce overthinking 
caused by \emph{cognitive overload} during navigation.
Specifically, we focus on exploiting the spatiotemporal localities unique to VLN tasks:
(1) The spatial locality:
In a panorama, 
we find that navigable views and a few neighboring views 
are critical for successful navigation.
We design a weighting mechanism 
that significantly reduces the number of views 
the encoder should process.
We also develop an efficient subgoal module 
that predicts navigable views from laser scans,
enabling compatibility with continuous environments
where such views are unknown to the agent as priors.
(2) The temporal locality:
We find that an agent encounters
identical or nearly identical views
in consecutive navigation steps.
We design a locality-sensitive hashing algorithm 
to avoid computing these matching views during navigation.
(3) %
We lastly develop an algorithm
for dynamically adapting the thresholds 
for an existing early-exit method based on the locality
to further reduce computations.

\emph{Third}, we comprehensively evaluate our input-adaptive navigation method
on 7 VLN benchmarks across three popular agents.
Our method reduces computations by up to 60\% with an average drop in SR of 11.7\% in the standard setting. In the more challenging continuous setting, 
it achieves $\sim$86\% savings with an even smaller 8\% SR decrease.
In contrast, baseline methods experience 
up to 33.6\% performance loss and fail to reduce computations.
Our ablation study also shows how a practitioner 
can configure our method for their navigation environments 
and the factors we do not rely on.
Moreover, we examine the robustness of our method 
to natural visual corruptions that may occur 
during navigation (such as lighting changes).
We show that while both the baseline and our method show 
a slight increase in the computations, 
our approach loses 7--10\% more performance.

\section{Related Work}
\label{sec:prelim}

\topic{Vision-and-language navigation (VLN).}
Research in this area has been supported by the development of high-quality simulators 
such as Matterport3D~\cite{Matterport3D} and Habitat~\cite{savva2019habitat}, which we leverage in our work.
Agents developed towards this challenging problem have ranged from earlier
recurrent models~\cite{R2R, fried2018speaker} to more recent transformer-based models~\cite{vlnbert, HAMT, chen2022think, wang2023scaling, kamath2023newpath, li2023improving, perincherry2025visual}. 
While recent agents achieve superior 
performance, 
larger models combined with high-fidelity panoramic observation and action spaces have led to
their increased complexity and higher computational costs during inference.
Our work is the first 
providing a tunable trade-off between computational demands and accuracy. 

VLN agents are studied in two environmental settings. 
The first is discrete (standard) VLN, where agents teleport between neighboring nodes in a known navigation graph that provides candidate navigable views as an agent's action space. 
To circumvent the unrealistic assumptions of navigation graphs, prior work~\cite{krantz2020vlnce} proposes continuous VLN, where agents instead use low-level actions and estimate navigable views using a sub-goal generation module. 
We empirically achieve large computational savings in both settings.

\topic{Input-adaptive mechanisms for computational efficiency.}
Prior work introduces two distinct mechanisms for input-adaptive inference:
adaptive neural networks (AdNNs) and multi-exit architectures.
AdNNs~\cite{wang2018skipnet, figurnov2017spatially} dynamically skip
certain blocks of the model to save computations during inference.
In contrast, multi-exit architectures~\cite{teerapittayanon2016branchynet,msdnet,kaya2019shallow,xin2020deebert} 
introduce an additional component to the model, 
such as classifiers attached to each internal layer (early-exits),
allowing the model to preemptively stop running forwards 
once stopping criteria are met.
Both mechanisms demonstrate computational savings 
while minimizing performance loss in classification tasks
(e.g., a 50\% reduction in computation at a utility loss of $\sim$10\%).
We use multi-exit architectures, as AdNNs are limited to residual networks.
Most multi-exit architectures are developed for classification tasks
and are \emph{not} compatible with VLN,
where an agent utilizes visual and/or language representations
generated from encoders.
The closest work by Tang~\textit{et al.}~\cite{MuE} developed 
an adaptation (MuE) to Transformer-based encoders, but despite our best efforts, 
it does not provide any computational savings in VLN tasks
(shown in Sec~\ref{subsec:overthinking-in-vln}).
Similarly, Yue \textit{et al.}~\cite{yue2024deer} propose an early-exit strategy for MLLM-based embodied AI, but do not address navigation tasks and focus on sequential action predictions rather than spatio-temporal dependencies in visual observations.
A separate line of research
explores methods for compressing models,
such as quantization and pruning.
These methods are orthogonal to our study
and can be applied in conjunction with \Ourmethod{}
(see Appendix~\ref{appendix:model-compression}).

\begin{figure*}[t]
\centering
\vspace{-0.1em}
\includegraphics[width=\linewidth]{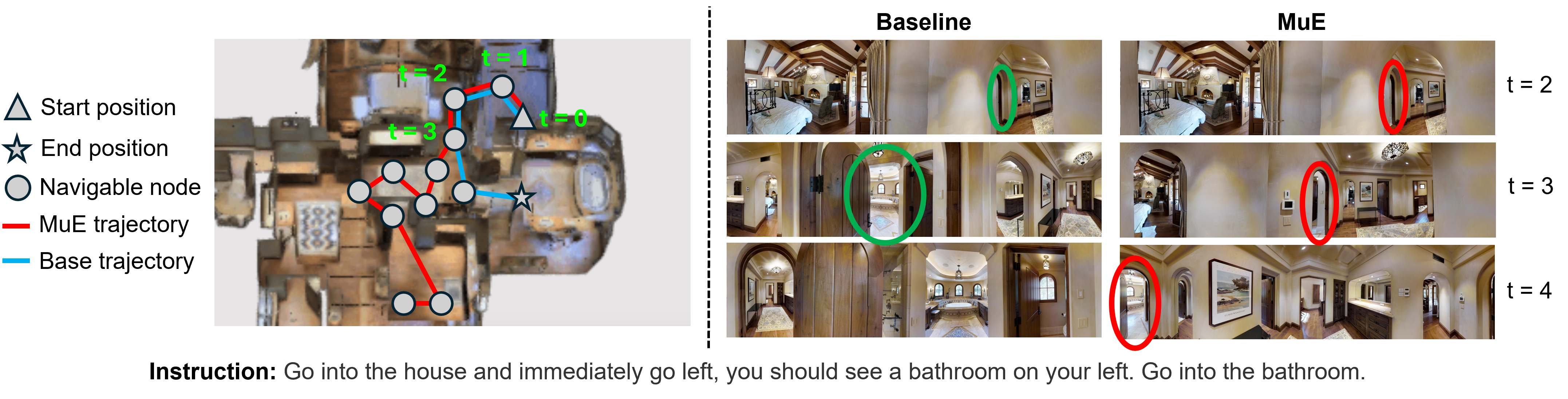}
\vspace{-2em}
\caption{%
    \textbf{Problems in employing existing input-adaptive methods in VLN.}
    We show that employing existing strategies leads to performance loss and an increase in computations. 
    (Left) The increase in computations stems from inappropriate navigation actions,
    and (Right) such decisions come from the inaccurate understanding of the visual world,
    e.g., the agent confuses where to navigate.}
\label{fig:challenges}
\vspace{-1em}
\end{figure*}

\section{Input-Adaptive Efficient VLN}
\label{sec:method}

\subsection{Characterizing Overthinking in VLN}
\label{subsec:overthinking-in-vln}

\topic{Computational bottleneck.}
The first step in designing an efficient input-adaptive mechanism
is to understand the computational bottleneck of an agent during navigation.
Because no prior work has studied
which component consumes the most computational resources,
we identify the bottleneck by analyzing the GFLOPs of each component in HAMT using the pre-trained agent on the R2R validation (unseen) set.

\begin{table}[h]
\centering
\adjustbox{max width=0.8\linewidth}{
    \begin{tabular}{@{}ccccc@{}}
    \toprule
     & \textbf{ViT} & \textbf{BERT} & \textbf{H-ViT} & \textbf{CMT} \\ \midrule
    \multicolumn{1}{r}{\textbf{GFLOPs (\%)}} & 99.50\% & 0.04\% & 0.07\% & 0.39\% \\ \bottomrule
    \end{tabular}
}
\vspace{-0.5em}
\caption{%
    \textbf{Component-wise computational demands.} 
    We run HAMT on the validation (unseen) set of R2R.}
\label{fig:bottleneck}
\vspace{-0.5em}
\end{table}

Table~\ref{fig:bottleneck} summarizes our result.
BERT requires the least computations (0.04\%) as it is used only once at the beginning of navigation to encode the human instruction. In contrast, 99.5\% of the computations come from the ViT, which must process 36 views per panorama at each navigation step.
Considering that the remaining components, H-ViT and CMT, only account for 0.46\% of the total computations, we decide to focus on the visual encoder.
We note that while DUET and \vlncebert have different architectures, the image encoder poses a similar bottleneck of at least 99.5\%.

\topic{Existing mechanisms are ineffective for VLN.}
Next, we examine whether existing input-adaptive inference methods 
can provide computational savings in VLN.
We find that most approaches
discussed in Sec~\ref{sec:prelim}
are incompatible with VLN settings 
because they are designed for classification tasks and not encoder models.
Tang~\textit{et al.}~\cite{MuE} proposes 
an input-adaptive strategy, MuE, tailored for encoder models.
MuE measures the cosine similarity between the output activations
from two consecutive transformer layers
to determine when to stop a forward pass.
If the cosine similarity becomes greater than a predefined threshold,
MuE stops forwarding and skips subsequent layers.
We test the MuE strategy on the ViT model in HAMT
and evaluate the performance and \ourmetric{} of the agent
on the validation (unseen) set of R2R.

\begin{table}[t]
\centering
\adjustbox{max width=\linewidth}{
\begin{tabular}{lccccc@{}}
\toprule
 \multirow{2.4}{*}{\textbf{Method}} & \multicolumn{4}{c}{\textbf{Performance}} & \multirow{2.4}{*}{\textbf{\Ourmetric{}}($\downarrow$)} \\ \cmidrule{2-5}
 & \textbf{TL}($\downarrow$) & \textbf{OSR}($\uparrow$) & \textbf{SR}($\uparrow$) & \textbf{SPL}($\uparrow$) & \\ \midrule \midrule
 Base& 11.53& 74.29& 66.16& 61.49& 4763.24 \\ 
 MuE & 17.37& 62.20& 43.93& 36.92& 4409.62 \\ \bottomrule
\end{tabular}
}
\vspace{-0.5em}
\caption{%
    \textbf{Performance and computational savings in HAMT with MuE.}
    Our adaptation of MuE leads to only marginal computational savings 
    at the cost of significant performance degradation.}
\label{tbl:MuE_baseline_comparison}
\vspace{-1em}
\end{table}

\topic{Results.}
Table~\ref{tbl:MuE_baseline_comparison} shows our results for the original and MuE-based HAMT agents. With an early-exit threshold of 0.998 (optimized for the best performance-efficiency trade-off, see Appendix~\ref{appendix:mue-with-diff-thresholds}), MuE reduces GFLOPs by 7\% but significantly degrades performance (up to 40\%).
The average GFLOPs per step with MuE is 406.10, compared to 607.06 in the baseline.
However, despite the significant per step GFLOPs savings, the total GFLOPs per trajectory increases because the MuE agent takes more steps to complete each trajectory.

In Figure~\ref{fig:challenges},
we analyze the factors contributing to 
the performance loss and the limited computational savings.
The left figure compares the trajectories of the original HAMT agent and HAMT with MuE.
Both agents navigate to the same position until $t$ = 2.
At $t$ = 3, the original HAMT agent correctly identifies the bathroom (green circle, top-right) and navigates to its front.
However, the MuE agent only takes a small step forward and then continues to make incorrect steps until reaching the step limit.
For MuE, processing fewer transformer layers led to 
an inaccurate understanding of the visual surroundings.
As shown in the bottom-right figures, the bathroom remains visible across steps ($t \in [2, 10]$), yet the MuE agent fails to recognize it and makes suboptimal decisions.
Appendix~\ref{appendix:mue-with-diff-thresholds} 
provides a further discussion on why MuE fails when applied 
directly to VLN.

\subsection{Our Methodology}
\label{subsec:our-method}

Prior work on input-adaptive inference treats each input independently.
As a result, existing methods inherit the \emph{one-size-fits-all} philosophy: 
a model adopts a single set of configurations, 
such as the early-exit threshold,
for all inputs. 
However, in dynamic settings, such as an agent navigating the physical world, inputs are not independent and depend on each other both spatially and temporally.

We introduce a novel input-adaptive inference method harnessing this unique property---spatial and temporal dependencies in the input. 
We first leverage spatial locality (Sec~\ref{subsubsec:harnessing-spatial-locality}): 
among the 36 views observed by an agent at each step, 
we find that those close to \emph{navigable views}---views the agent can navigate to---are important.
We then propose a novel approach to assign the exit thresholds of 
an existing input-adaptive inference method (Sec~\ref{subsubsec:adaptation-of-mue})
for the non-masked views to provide further computational savings.
In Sec~\ref{subsubsec:harnessing-temporal-locality},
we exploit temporal locality:
between panorama views observed across steps,
most views overlap and do not require their forward passes to be run again.
Finally, while our methods are directly applicable to standard VLN agents that navigate predefined traversal graphs, we also extend them to the more practical continuous setting in Sec~\ref{subsubsec:input-adaptive-continuous}.

\subsubsection{Harnessing spatial locality}
\label{subsubsec:harnessing-spatial-locality}

Each panorama has 36 views, and the agent computes visual embeddings for each view at every navigation step.
We hypothesize that only \emph{navigable views} are crucial for navigation.
Intuitively, these views form the agent's decision space, so the information they contain should suffice for choosing the proper action.
To test this hypothesis, we retain all navigable views and mask the remaining views (setting them to zero).
This prevents the ViT from processing masked views, reducing computation.
We evaluate the effectiveness of this approach 
with the HAMT agent on the validation (unseen) set of R2R.
We find that it results in an 84\% gain in efficiency but at the cost of a 33\% reduction in SR.

\begin{figure}[h]
\centering
\includegraphics[width=1.0\linewidth]{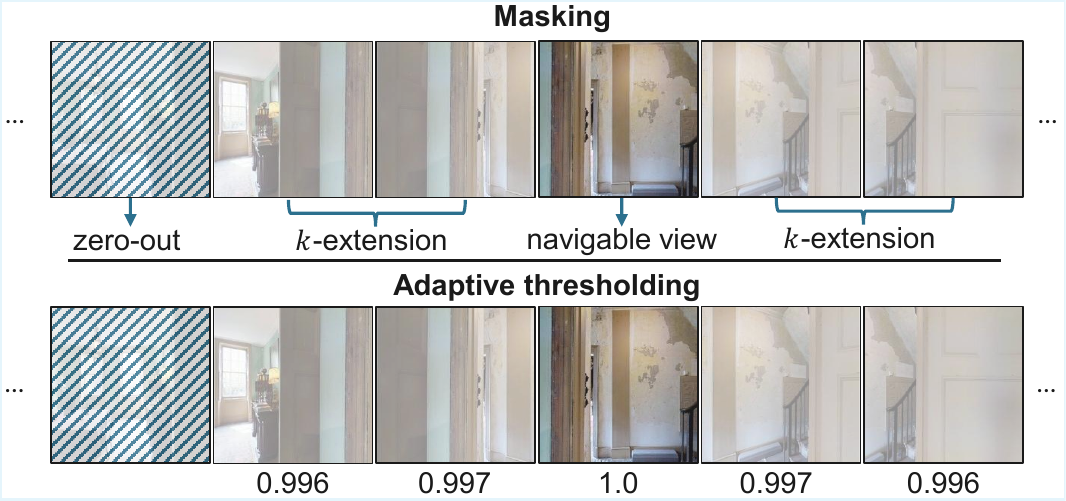}
\vspace{-2em}
\caption{%
    \textbf{Our masking and thresholding.}
    The top figure shows how we mask non-navigable views,
    and the bottom figure shows how we adaptively assign the exit thresholds of MuE.}
\label{fig:reason-for-failure}
\end{figure}

To understand this issue, we analyze cases where masking non-navigable views prevents the agent from reaching the target.
In Figure~\ref{fig:reason-for-failure}, processing only the navigable view (\nth{4} from the left) may obscure whether the path leads to a stairway.
However, processing neighboring views increases the likelihood of correctly recognizing the path.

\topic{$k$-extension.}
To address this, 
we extend the number of views the agent processes near the navigable views by $k$.
Let $V$ be the set of $n$ navigable views, with each $v_i$ indexed by $\{1, 2, \dots, n\}$.
The $k$-extension $V_k^i$ for navigable view $i$ is:
\[V_k^{i} = \{v_i^j|max(1, i - k) \,\leq j \leq min(i + k, 36) \,\},\]
where $v_i^j$ is a non-navigable view.
The union of $V_k^i$'s gives the views to process, leaving $36-|V_k|$ masked.
With a careful calibration of $k$,
we reduce the total computations by 2$\times$ times 
while keeping the performance drop near 10\%.
In our evaluation, setting $k=4$--$6$ offers the best trade-off.

\subsubsection{Using adaptive thresholds as stopping criteria}
\label{subsubsec:adaptation-of-mue}

On top of our $k$-extension,
we design an adaptive mechanism to early-exit extended views and further improve the speed-up.
As described in previous sections,
we focus on MuE, the only early-exit mechanism 
compatible with encoder models.

\topic{Using budgeted-batch inference.}
The current implementation of MuE processes 
each test sample with its input-adaptive mechanism.
However, this \emph{per-sample, anytime} strategy 
is incompatible with our scenario, 
where the agent processes a batch of 36 views 
in a single panorama at once with the ViT.
While each view in the batch
should ideally exit at different layers, 
this per-sample approach forces all the views to use the same exit layer.
To address this issue, we employ 
\emph{budgeted-batch inference}~\cite{msdnet}: each sample in a batch
uses ``uneven" computations, meaning that
processing can stop at different layers for each sample, all within a set computational budget.
We assign a sufficiently large budget
so that the mechanism can handle the worst-case complexity,
where none of the samples utilize early stopping.

\topic{Our adaptive thresholding.}
In Sec~\ref{subsubsec:harnessing-spatial-locality},
we find that navigable views are most important, 
and a view’s importance decreases with distance from navigable views.
We thus design a mechanism to apply early-exit thresholds 
differently based on the importance of each view.
We propose a concept, \emph{rank}: low-rank views receive an aggressive (larger) threshold, while high-ranked views receive a conservative (smaller) threshold.
Suppose we have a navigable view $v_i$ 
at index $i$ in a panorama and
$k$ is the number of extended views near $v_i$.
We define the rank $R_{i,j}$ of a non-navigable view $v_j$ relative to $v_i$ as the difference between the indices $|j - i|$.
We do not process when $R_{i,j} \geq k$,
as views beyond $k$ are masked.
We still fully process the navigable views to retain performance.
We then assign the exit threshold $T_{i,j}$
(the cosine similarity) for MuE as follows:
{\[T_{i,j} = T_0 \cdot e^{(-A \cdot R_{i,j})}\]}%
where $T_0$ is the initial threshold set to 1.0,
$A$ is the aggressiveness we set to $9 \times 10^{-4}$, and
$R_{i,j}$ is the rank computed above.
Note that the threshold decreases as the rank increases.

\subsubsection{Harnessing temporal locality}
\label{subsubsec:harnessing-temporal-locality}

Our final insight is that during navigation,
an agent will encounter similar views multiple times,
leading to \emph{temporal redundancy}.
For example, views at step $i$ 
are similar to those at step $i+1$.
The agent may also revisit the same surroundings
due to misleading navigation or encounter similar  
but less important surroundings,
such as ceilings or walls.

To reduce temporal redundancy,
we employ locality-sensitive hashing (LSH) to store and retrieve similar visual representations, avoiding redundant processing.
We use SimHash~\cite{charikar2002similarity, andoni2008near}, which maps high-dimensional RGB views to low-dimensional binary encodings via random projection.
Given a view $v$ and randomly initialized hyperplanes $\{h_i\}_{i \in \{1, \dots, n\}}$, the algorithm determines which side of the hyperplane $v$ falls on via the dot-product of $v$ and $h_i$. 
If $v$ is on the top side of $h_i$,
SimHash assigns 1; otherwise, it is 0.
Similar views are then encoded as
the same binary encoding of length $n$, e.g., $010\dots1$,
which we use as a key to store view-encoding pairs.
Views mapped to the same key are then reused if they are sufficiently similar, which we measure using their cosine similarity.
To balance performance and efficiency, we set $n$ to 10 and the similarity threshold to 0.85 and 0.95 for standard and continuous VLN, respectively.
Like early-exiting, we do not hash navigable views 
and fully process them.
Note that with our $k$-extension,
we limit the space complexity
of caching by
storing only a subset of views.
With this mechanism,
we achieve an additional 2--4\% computational savings with minimal utility loss.
See Appendix~\ref{appendix:rplsh-algorith-detail} for more details and storage overhead analysis.

\begin{figure}[ht]
\begin{minipage}{\linewidth}
\vspace{-1.2em}     %
\begin{algorithm}[H]
\caption{Our Input-adaptive Navigation at Each Step}
\label{alg:our-algorithm}
\textbf{Input:} 
    a panorama $P$, 
    navigable views $V$,
    visual encoder $f_{\theta}$,
    hash table $h$,
    and the number of views to extend $k$ \\
\textbf{Output:} 
    a set of visual representations $E$ for views in $P$
\begin{algorithmic}[1]
\State $E \gets \varnothing$
\For {$i = 1, 2, \ldots, 36$} \Comment{Iterate over views in $P$}
    \State 	$v_i \gets P[i]$
    \If{$v_i$ in $V$}
        \State $e_i \gets f_{\theta}(v_i)$
        \State $E \gets E + e_i$
    \ElsIf{$i$ in $k$ proximity of any views in $V$}
        \State $e_i \gets h(v_i)$
        \If{$e_i$ does not exist}
            \State $j \gets$ the index of the closest navigable view
            \State $T_i \gets $ \texttt{ComputeThreshold}($R_{i,j}$)
            \State $e_i \gets$ \texttt{RunMuEInference}($v_i, T_i$)
            \State $h \gets$ \texttt{AddToHashTable}($h, v_i, e_i$)
        \EndIf
        \State $E \gets E + e_i$
    \Else
        \State $E \gets E + \vec{\scriptstyle 0}$
    \EndIf
\EndFor
\State \Return $E$
\end{algorithmic}
\end{algorithm}
\vspace{-2.0em}     %
\end{minipage}
\end{figure}

\subsubsection{Input-adaptive inference for continuous VLN}
\label{subsubsec:input-adaptive-continuous}

In continuous VLN~\cite{krantz2020vlnce}, agents navigate 3D environments without predefined traversal graphs, requiring them to predict navigable views.
Existing agents~\cite{krantz2022sim, hong2022bridging} address this with \emph{subgoal generation modules} (SGMs), which process 2D laser occupancy scans and encoded images to predict navigable views. 
However, this conflicts with our input-adaptive inference techniques, as the entire panorama must be processed before identifying navigable views.

To solve this problem, we introduce a \emph{scan-only} SGM that predicts navigable views using only laser occupancy scans. 
We use the U-Net SGM from Krantz~\textit{et al.}~\cite{krantz2022sim}, but remove image feature processing.
Following their training procedure, we minimize the Sinkhorn divergence~\cite{cuturi2013sinkhorn} between predictions and ground-truth subgoals. 
Our scan-only SGM achieves a validation loss of 0.63 on Matterport3D scene data, the same as the original work.
With the ability to predict navigable views prior to image encoding, our methods become compatible with continuous VLN agents.

\subsection{Putting All Together}
\label{subsec:combine-them-all}

Now, we describe how our three mechanisms
are combined to perform input-adaptive inference on a panorama.
We show the pseudo-code of our method in Algorithm~\ref{alg:our-algorithm}:

\topic{(line 1-2) Initialize.}
It takes a panorama $P$ and returns  
the visual representations of its 36 component views.  
We initialize the output $E$ as empty and iterate over each view.

\topic{(line 4-6) Compute the representation of a navigable view.}
If the currently chosen view $v_i$ is a navigable view,
we fully compute its visual representation $e_i$ and add it to the set $E$.

\topic{(line 7-15) Retrieve (or compute) the representation of the extended views.}
In Sec~\ref{subsubsec:harnessing-spatial-locality},
to improve the visual understanding,
we develop the $k$-extension.
We process $k$ views on both sides (left/right) of a navigable view.
If $v_i$'s representation is in the hash table $h$, we retrieve $e_i$ and add it to $E$;
otherwise, we compute $e_i$.
Note that the hash table $h$ is initialized at the first step of the navigation.
To compute $e_i$, we determine $v_i$'s rank $R_{i,j}$ and 
decide the exit threshold $T_i$. %
We run the inference with ViT, adapted for MuE, using $T_i$
and store the output $e_i$ into $h$ and $E$.

\topic{(line 17) Skipping the masked view.}
If $v_i$ is neither a navigable view nor in its $k$-extension,
we store a zero-vector and move on to the next view $v_{i+1}$.

\section{Evaluation}
\label{sec:results}

\label{sec:exp-setup}

\topic{Datasets.}
Following prior work,
we evaluate standard VLN with six datasets:
Room-to-Room (R2R)~\cite{R2R}, R2R-Back~\cite{HAMT}, R2R-Last~\cite{HAMT},
REVERIE~\cite{qi2020reverie}, CVDN~\cite{thomason2020vision}, and SOON~\cite{zhu2021soon}.
For REVERIE, we set $k = 6$ for $k$-extensions to minimize performance loss while ensuring a roughly 50\% speed-up. For all other benchmarks, we achieve this using $k = 4$.
To evaluate continuous VLN, we use R2R-CE~\cite{krantz2020vlnce}. 

\topic{VLN agents.}
We evaluate two off-the-shelf standard VLN agents: HAMT~\cite{HAMT} and DUET~\cite{chen2022think}. 
HAMT uses a ViT~\cite{ViT} for vision, BERT~\cite{devlin-etal-2019-bert} for language, and a hierarchical ViT for temporal context, predicting actions via a cross-modal Transformer.
DUET also employs ViT and BERT but integrates object features (e.g., bounding boxes) and separates planning into global and local cross-modal encoders, fusing their outputs for action prediction.
For continuous VLN, we use \vlncebert~\cite{krantz2022sim}, which uses ResNet-152~\cite{he2016deep} for visual encoding and BERT for recurrently processing visual and language information and predicting actions.

\topic{Evaluation metrics.}
We evaluate navigation success using four metrics from the prior work~\cite{HAMT, krantz2022sim}:  
(1) Trajectory length (TL): path length of the agent in meters,
(2) oracle success rate (OSR): fraction of paths with at least one viewpoint within 3 meters of the target,
(3) success rate (SR): fraction of final positions within 3 meters of the target and,
(4) success rate normalized by inverse path length (SPL): SR normalized by
the ratio between the shortest path length and the predicted
path length.  
For computational efficiency, we measure the GFLOPs and wall time per navigation; however, we prioritize GFLOPs as wall time depends on hardware and software implementation (see Appendix~\ref{appendix:full-results} for details).

For REVERIE and SOON, additional object features are used for navigation.
We could not find the original feature extraction implementations, so we use cached object features and apply our strategy only to image feature extraction.
We then report the GFLOPs for image feature processing and treat the cost of object feature extraction as a constant ($C$).
All other benchmarks do not use object features.

\subsection{Effectiveness in the Standard VLN Setting}
\label{subsec:effectiveness}

We evaluate our method in standard VLN with two agents, six benchmarks, and five metrics described in Sec~\ref{sec:exp-setup}.
We compare with two baselines:
no input-adaptive methods (Base)
and MuE, adapted for each agent 
to provide the optimal performance-efficiency trade-off.
For \Ourmethod{}, we present four variations:
one with $k$-extension, two adding mechanisms ($+$LSH, $+$thresholds), and one combining all.

\begin{table}[h]
\centering
\adjustbox{max width=\linewidth}{
\begin{tabular}{@{}clccccc@{}}
\toprule
\multirow{2.5}{*}{\textbf{Agent}} & \multirow{2.5}{*}{\textbf{Method}} & \multicolumn{4}{c}{\textbf{Performance}} & \multirow{2.5}{*}{\textbf{\Ourmetric{}}($\downarrow$)} \\ \cmidrule(l){3-6} 
 &  & \textbf{TL}($\downarrow$) & \textbf{OSR}($\uparrow$)& \textbf{SR}($\uparrow$) & \textbf{SPL}($\uparrow$) &  \\ \midrule \midrule
\multirow{6}{*}{HAMT} & Base & 14.07 & 35.73 & 31.81 & 29.17 & 5434.71+$C$ \\ \cmidrule{2-7}
 & MuE & 18.13 & 22.92 & 13.83 & 10.10 & 4098.77+$C$ \\ \cmidrule{2-7}
 & Ours ($k$-extension) & 13.85 & \textbf{26.53} & \textbf{24.96} & \textbf{22.97} & 3121.20+$C$ \\
 & Ours ($k$-extension+LSH) & 13.84 & \textbf{26.53} & \textbf{24.96} & \textbf{22.97} & 2359.72+$C$ \\
 & Ours ($k$-extension+thresholds) & 13.25 & 26.44 & 24.60 & 22.82 & 2723.01+$C$  \\
 & Ours (All) & \textbf{13.22} & 26.47 & 24.62 & 22.85 & \textbf{2073.69}+$C$ \\ \midrule
 \multirow{6}{*}{DUET} & Base & 22.49 & 51.46 & 47.09 & 33.54 & 6185.15+$C$ \\ \cmidrule{2-7}
 & MuE & 32.65 & 33.23 & 27.35 & 15.93 & 4888.35+$C$ \\ \cmidrule{2-7}
 & Ours ($k$-extension) & \textbf{21.43} & 46.58 & 41.81 & 29.29 & 3674.29+$C$ \\
 & Ours ($k$-extension+LSH) & 21.44 & \textbf{46.75} & \textbf{41.95} & \textbf{29.48} & 3381.45+$C$ \\
 & Ours ($k$-extension+thresholds) & 22.79 & 44.96 & 39.28 & 27.00 & 3399.44+$C$ \\
 & Ours (All) & 22.81 & 45.07 & 39.36 & 27.14 & \textbf{3145.92}+$C$ \\ \bottomrule
\end{tabular}
}
\vspace{-0.5em}
\caption{%
    \textbf{Effectiveness of our input-adaptive inference method for standard VLN.}
    We show our results on REVERIE for the HAMT and DUET agents.
    Each cell contains the averaged metric over the trajectories in the validation (unseen) set. 
    $C$ is the constant cost of object feature extraction.
    For each metric and model, the best result across the input-adaptive methods is \textbf{bolded}.
    }
\label{tbl:main-results}
\vspace{-0.5em}
\end{table}

\topic{Results.}
Table~\ref{tbl:main-results} summarizes our results for REVERIE, which we prioritize because it is generally more challenging than other benchmarks~\cite{HAMT, chen2022think}.
Due to the page limit,
we show the full results for other benchmarks in Appendix~\ref{appendix:full-results} and more combinations of mechanisms in Appendix~\ref{appendix:per-mech-analysis}. 

For REVERIE, applying all of our mechanisms 
saves 49--62\% computation (excluding object features)
while maintaining an SR loss between 16.4--22.6\%;
across all benchmarks (see Appendix~\ref{appendix:full-results}), the average reduction in computations is 56\% with just a 11.7\% drop in SR.
We set the upper limit for performance loss near 10--20\% for most tasks,
consistent with prior work on input-adaptive inference methods%
~\cite{msdnet, liu2020fastbert, kaya2019shallow, xin2020deebert, MuE}.
The naive adaptations of MuE only provide 21.0--24.6\% computational savings
and experience a significant performance drop of 41.9--56.5\% in SR,
as expected from our initial investigation in Sec~\ref{subsec:overthinking-in-vln}.
Our $k$-extension alone provides 
a 40.6--42.6\% reduction in \ourmetric{}
with only a 11.2--21.5\% drop in SR.
If we apply the adaptive thresholding ($+$thresholds), 
we achieve an additional 7.5--12.8\% computational savings, 
with a marginal performance loss of $\sim$1--6\%.
Separately, combining the LSH with the $k$-extension
results in additional computational savings up to 24.4\%, 
with no performance loss (the SR even increases).

\begin{table}[ht]
\centering
\adjustbox{max width=\linewidth}{
\begin{tabular}{lccccc}
\toprule
\multicolumn{1}{l}{\multirow{2}{*}{\textbf{Method}}} & \multicolumn{4}{c}{\textbf{Performance}}                                                 & \multirow{2}{*}{\textbf{GFLOPs}($\downarrow$)} \\ \cmidrule{2-5}
\multicolumn{1}{c}{}                        & \multicolumn{1}{l}{\textbf{TL}($\downarrow$)} & \multicolumn{1}{l}{\textbf{OSR}($\uparrow$)} & \multicolumn{1}{l}{\textbf{SR}($\uparrow$)} & \textbf{SPL}($\uparrow$) &                         \\ \midrule \midrule
Base                                        & 10.59                  & 51.88                   & 43.23                  & 36.53 & 18074.05                \\ \midrule
SGM                               & 9.79                   & 46.82                   & 39.86                  & 34.76 & 2396.54                 \\
SGM+LSH                           & 10.32                  & 45.79                   & 37.14                  & 31.95 & 1741.33                 \\ \bottomrule        
\end{tabular}
}
\vspace{-0.5em}
\caption{\textbf{Continuous VLN results.} The performance and computational savings for the baseline and our efficient \vlncebert agents on the R2R-CE validation (unseen) set.}
\label{tbl:continuous-results}
\vspace{-1em}
\end{table}

\subsection{Effectiveness in Continuous Environments}
\label{subsec:continuous-results}

We now study the effectiveness of our method in continuous VLN.  
Unlike standard agents, \vlncebert determines actions using only navigable views. With our scan-only SGM, this allows the agent to entirely disregard non-navigable views, eliminating the need for $k$-extensions.
Additionally, ResNet is incompatible with MuE, preventing the use of our early-exit strategy.
Therefore, we evaluate two input-adaptive variants alongside the base agent: one using our scan-only SGM and the other combining it with LSH.

\topic{Results.} Table~\ref{tbl:continuous-results} shows our results on R2R-CE. 
We first find that the baseline agent requires substantially more computations than agents in the standard VLN setting. 
This is primarily because viewpoints are higher resolution ($3 \times 480 \times 640$ versus $3 \times 224 \times 224$), therefore requiring more GFLOPs to process through the visual encoder. 
Performance is also lower, as R2R-CE is far more challenging than its discrete counterpart~\cite{krantz2022sim}. 
Despite this, our proposed techniques offer large computational savings with minimal performance drop.
When applying our scan-only SGM (SGM) to \vlncebert, we achieve an 87\% reduction in GFLOPs while SR only drops by 8\%. 
By predicting the navigable viewpoints \emph{before} encoding them, our agent adaptively processes only the navigable views instead of the entire panorama. 
This results in just 5 out of 36 views being processed per navigation step, on average. 
Computations are reduced by $\sim$90\% by incorporating LSH (+LSH);
however, as we find in Sec~\ref{subsubsec:harnessing-spatial-locality}, navigable views are more critical to navigation, so caching them leads to a larger SR drop of 14\%.

\subsection{Sensitivity to Our Method's Configurations}
\label{subsec:ablation}

Next, we assess the sensitivity of our method's computational savings to its configurations.
Our method's effectiveness depends on three key configurations: 
the number of extended views ($k$), 
the adaptive thresholds set based on the extension, 
and the similarity measure used in our LSH mechanism.
Here, we show our results for R2R.

\begin{table}[ht]
\centering
\adjustbox{max width=0.8\linewidth}{
\begin{tabular}{c|cccc|c} 
    \toprule
    \multirow{2.5}{*}{\textbf{$k$}} & \multicolumn{4}{c|}{\textbf{Performance}}& \multirow{2.5}{*}{\textbf{\Ourmetric{}}($\downarrow$)} \\ \cmidrule(l){2-5}
     &  \textbf{TL}($\downarrow$) &  \textbf{OSR}($\uparrow$)& \textbf{SR}($\uparrow$) &  \textbf{SPL}($\uparrow$) &  \\ \midrule \midrule
     - & 11.53& 74.29& 66.16& 61.49& 4763.24 \\ \midrule
     1 & 15.38& 70.20& 54.32& 46.96& 1250.65 \\ %
     2 & 13.67& 70.84& 58.19& 51.99& 1554.82 \\ %
     3 & 12.94& 71.60& 60.20& 54.60& 1793.76 \\ %
     4 & 12.52& 71.90& 61.17& 55.63& 2013.48 \\ %
     5 & 12.19& 71.99& 62.32& 57.08& 2216.34 \\ %
     6 & 11.89& 71.99& 62.84& 57.94& 2414.46 \\ %
     \bottomrule
\end{tabular}
}
\vspace{-0.5em}
\caption{%
\textbf{Performance and computational savings across different $k$ values.}
We evaluate with the HAMT agent in R2R. 
}
\label{tbl:ablation-extension}
\vspace{-1em}
\end{table}

\topic{Number of extended views $k$.}
Table~\ref{tbl:ablation-extension} shows performance and \ourmetric{} across $k \in [1, 6]$.
As $k$ decreases, 
the agent processes fewer views in each panorama,
yielding 49–74\% computational savings at a 5–18\% performance cost.
Surprisingly, with $k=1$, 
we save 74\% of GFLOPs 
while only sacrificing 18\% in performance (SR).
We choose $k$ such that an agent processes approximately half of the total views in each panorama; this results in $k=4$--$6$ for the benchmarks we consider.
Given that this strategy provides 50\% computational savings
across all benchmarks, even when the average number of navigable views 
per panorama is not used to set $k$,
we believe the strategy is transferable to new settings.

\begin{table}[t]
\centering
\adjustbox{max width=\linewidth}{%
\begin{tabular}{c|cccc||cccc|c} 
    \toprule
     & \multicolumn{4}{c||}{\textbf{Thresholds $T$}}&  \multicolumn{4}{c|}{\textbf{Performance}} & \\
     $A$ & $R_{1,j}$ & $R_{2,j}$ & $R_{3,j}$ & $R_{4,j}$ & \textbf{TL}($\downarrow$) & \textbf{OSR}($\uparrow$)& \textbf{SR}($\uparrow$) & \textbf{SPL}($\uparrow$) & \textbf{\Ourmetric{}($\downarrow$)} \\ \midrule \midrule 
     0 & 1.0&  1.0&  1.0&  1.0&  12.52&  71.90&  61.17&  55.63&  2013.48 \\ %
     0.007 & 1.0&  1.0&  1.0&  0.997&  12.57&  71.60&  60.96&  55.32&  1973.23 \\ %
     0.009 & 1.0&  1.0&  0.997&  0.996&  12.87&  71.95&  60.41&  54.5&  1917.61 \\ %
     0.015 & 1.0&  0.997&  0.996&  0.993&  13.44&  70.67&  57.98&  52.09&  1848.89 \\ %
     0.022 & 0.997& 0.996& 0.993& 0.990& 14.61& 70.29& 55.60& 48.56& 1768.85 \\ \bottomrule %
\end{tabular}
}
\vspace{-0.5em}
\caption{%
\textbf{Performance and computational savings across different early-exit thresholds.}
We set the aggressiveness $A$ within [0.0, 0.022].
Note that we round the threshold to 3 decimal places
and set any thresholds greater than 0.998 to 1.0 
as ViTs with these thresholds will use full computations.}
\label{tbl:ablation_threshold}
\vspace{-1em}
\end{table}

\begin{figure*}[t]
	\centering
	\includegraphics[width=0.95\linewidth]{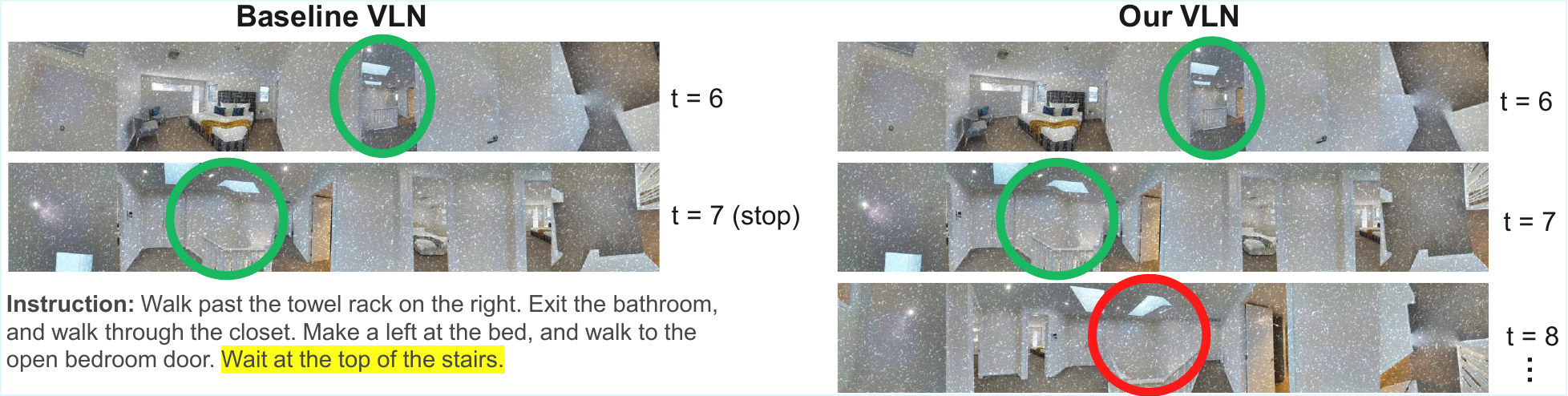}
	\vspace{-0.8em}
	\caption{%
		\textbf{Comparison of baseline and our agent trajectories under Spatter corruption.}
		We demonstrate that our agent fails to stop at the target location, resulting in incorrect navigation (Right), whereas the baseline agent successfully stops as instructed (Left).}
	\label{fig:spatter_trajectory}
	\vspace{-1em}
\end{figure*}

\topic{Early-exit thresholds.}
We also analyze the impact of the early-exit threshold $T$
by varying the aggressiveness factor $A$ from 0.0 to 0.0022;
the threshold decreases
as a view becomes farther from a navigable view.
Table~\ref{tbl:ablation_threshold}  shows that
increasing aggressiveness 
improves computational efficiency but reduces performance.
Using $A > 0.009$ causes an SR drop of over 10\%, so we set $A = 0.0009$.

\topic{Using different similarity metrics.}
In Sec~\ref{subsubsec:harnessing-temporal-locality},
our primary metric for computing similarity between views
is cosine similarity.
We explore whether employing different similarity metrics
can further enhance the effectiveness of our method.
To evaluate this, we test four additional metrics: 
visual features extracted from ViT's first-layer activations,
SSIM~\cite{wang2004image}, FSIM~\cite{zhang2011fsim}, and LPIPS~\cite{zhang2018unreasonable}.
We also test SURF~\cite{bay2006surf} and SIFT~\cite{lowe2004distinctive} in Appendix~\ref{appendix:similarity-metrics},
but they fail to match visually similar views in consecutive navigation steps.

\begin{table}[ht]
\vspace{-0.2em}     %
\centering
\adjustbox{max width=\linewidth}{
\begin{tabular}{@{}lccccc@{}}
\toprule
\multirow{2.4}{*}{\textbf{Similarity Metrics}} & \multicolumn{4}{c}{\textbf{Performance}} & \multirow{2.4}{*}{\textbf{\Ourmetric{}}($\downarrow$)} \\ \cmidrule(l){2-5}
& \textbf{TL}($\downarrow$) & \textbf{OSR}($\uparrow$)& \textbf{SR}($\uparrow$) & \textbf{SPL}($\uparrow$) & \\ \midrule \midrule
\multirow{1}{*}{\textbf{Cosine similarity (Ours)}}& 12.87 & 71.95& 60.41 & 54.50 & 1917.61 \\ \midrule
\multirow{1}{*}{\textbf{ViT (\nth{1} layer activation)}}& 12.89& 71.99& 60.41 & 54.59 & 1966.95 \\
\multirow{1}{*}{\textbf{SSIM}~\cite{wang2004image}}& 12.87& 71.95& 60.41& 54.57&1934.48 \\
\multirow{1}{*}{\textbf{FSIM}~\cite{zhang2011fsim}}& 12.88& 71.95& 60.45& 54.58&1937.73 \\
\multirow{1}{*}{\textbf{LPIPS}~\cite{zhang2018unreasonable}}& 12.87& 71.95& 60.49& 54.62&1925.15 \\ 
\bottomrule
\end{tabular}
}
\vspace{-0.5em}
\caption{%
    \textbf{Impact of employing different similarity metrics in LSH.}
    We experiment with the HAMT model in R2R.}
\label{tbl:ablation_different_sim_metric}
\vspace{-0.2em}     %
\end{table}

Table~\ref{tbl:ablation_different_sim_metric} shows our results.
Across the board, we observe only a marginal difference between the similarity metrics.
We see a performance increase of 0.16--0.22\%
at the cost of a 2.6\% increase in computation.
The largest increase in computation comes from 
obtaining the intermediate activation from ViT.
The results indicate that our method is not dependent on
the choice of similarity metrics, studied so far in prior work.
We also manually analyzed views deemed similar by these metrics, finding most to be identical or having slight variations, e.g., plain walls with lighting differences.

\begin{figure}[h]
\centering
\includegraphics[width=\linewidth]{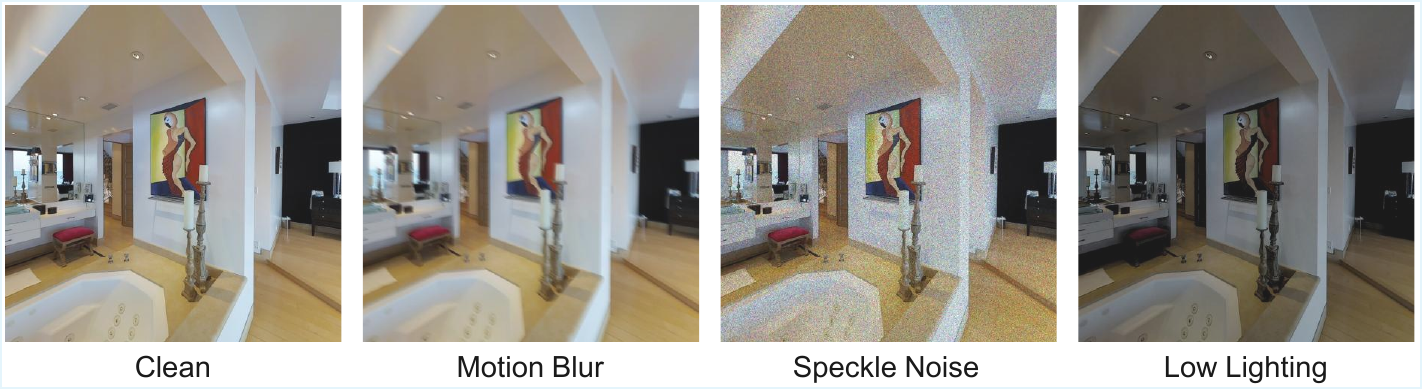}
\vspace{-2em}
\caption{%
    \textbf{Examples of the visual corruptions we consider.}}
\label{fig:visual_corruption}
\vspace{-0.5em}
\end{figure}

\subsection{Robustness to Natural Visual Corruptions}
\label{subsec:robustness}

Following recent work~\cite{chattopadhyay2021robustnav},
we evaluate the robustness of our method's efficiency to practical visual corruptions: Spatter, Defocus Blur, Speckle Noise, Low Lighting, and Motion Blur.
Figure~\ref{fig:visual_corruption} shows 
an example of the most distinct ones.
We apply each corruption to 
the entire validation (unseen) set of R2R, 
using the corruption framework by Chattopadhyay~\textit{et al.}~\cite{chattopadhyay2021robustnav}.
We set the severity to 3 out of 5,
because setting it above 3 
causes excessive distortion to the views,
which does not reflect the realistic corruptions 
an agent would encounter.

\begin{table}[h]
\centering
\adjustbox{max width=\linewidth}{
\begin{tabular}{clccccc@{}}
    \toprule
    \multirow{2}{*}{\textbf{Agent}} & \multirow{2}{*}{\textbf{Corruption}}& \multicolumn{4}{c}{\textbf{Performance}} & \\ 
     &  & \textbf{TL}($\downarrow$) & \textbf{OSR}($\uparrow$)& \textbf{SR}($\uparrow$) & \textbf{SPL}($\uparrow$) & \textbf{\Ourmetric{}}($\downarrow$) \\ \midrule \midrule
    \multirow{6}{*}{HAMT} & None & 11.53& 74.29& 66.16& 61.49& 4763.24 \\ \cmidrule{2-7}
     & Spatter& 13.30& 69.82& 58.71& 52.91& 5227.36\\    
     & Defocus Blur& 13.87& 66.50& 55.21& 49.32& 5383.35\\
     & Speckle Noise& 13.60 & 62.88& 51.68 & 46.02 & 5345.07 \\
     & Low Lighting& 12.15 & 71.31& 62.58 & 57.23 & 4903.06 \\
     & Motion Blur& 12.41 & 68.20& 59.13 & 54.01 & 4996.64 \\ \midrule
    \multirow{6}{*}{Ours}& None & 12.87& 71.95& 60.41& 54.50& {1917.61} \\ \cmidrule{2-7}
     & Spatter& 16.09& 67.01& 49.04& 41.53& 2201.19\\   
     & Defocus Blur& 16.22& 63.69& 49.21& 41.73& 2082.57\\
     & Speckle Noise& 18.11 & 61.43& 40.87 & 33.60 & 2342.67 \\
     & Low Lighting& 15.27 & 69.90& 52.58 & 45.33 & 1516.50 \\
     & Motion Blur& 14.47 & 65.47& 52.96 & 46.52 & 1986.50 \\ \bottomrule
\end{tabular}
}
\vspace{-0.5em}
\caption{%
    \textbf{Robustness evaluation of baseline HAMT and efficient HAMT under visual corruptions.} 
    We evaluate both models on R2R under clean conditions and five types of visual corruption.
    }
\label{tbl:OOD}
\vspace{-1.0em}
\end{table}

\topic{Results.}
Table~\ref{tbl:OOD} summarizes our findings
from evaluating the HAMT agent on the R2R benchmark. 
We first observe that applying our method to a VLN agent
reduces its performance and computational savings 
compared to the original agent.
Across the five corruptions,
the HAMT agent shows 5.4--21.1\% reductions in performance,
while our agent undergoes 12.3--31.3\% reductions.
GFLOPs increase by 2.9--13.0\% in HAMT,
while we show an increase of 3.6--20.9\%.
Per corruption, we find that 
both agents are most resilient to Low Lighting 
and least robust to Speckle Noise.
This aligns with the findings of prior work~\cite{chattopadhyay2021robustnav}.
HAMT and DUET use visual encoders pre-trained on ImageNet-1K, 
meaning they inherit the susceptibility 
of these ImageNet encoders to visual corruptions.
This finding highlights the importance of studies
on enhancing the robustness of visual encoders to natural corruptions%
~\cite{hendrycks2018benchmarking, guo2023improving, pmlr-v202-zhu23a}, 
which could improve the robustness across various VLN agents.

Interestingly, while our agent experiences a large drop in SR, the impact on OSR is notably smaller.
To uncover why, we manually analyze R2R trajectories.
Figure~\ref{fig:spatter_trajectory} shows a representative path for both agents.
Our agent consistently overshoots the target, whereas the baseline stops correctly.
The agent should stop at the top of the stairs, but instead moves past them into an adjacent room and continues turning until reaching the step limit.
This indicates that our agent can navigate to the target, but struggles to stop in the presence of visual corruption;
thus, we hypothesize that our approach mainly affects \emph{recognition} rather than navigation itself. 

To test improving the robustness, we apply a median filter (kernel size of 5) to denoise corrupted images.
On the most impactful corruption Speckle Noise, we recover SR by 17.9\% and reduce GFLOPs by 6.1\%.
This indicates that denoising is a promising direction for enhancing performance in corrupted environments; as robustness is a separate area of research, we leave further investigation as future work.

\section{Conclusion}
\label{sec:conclusion}

We propose an input-adaptive inference method 
to mitigate overthinking in vision-and-language navigation (VLN)
and achieve computational efficiency.
Unlike the overthinking problem in conventional domains, 
such as object recognition or natural language comprehension,
addressing overthinking in VLN presents three unique challenges:
(1) How can we leverage spatial locality in views
observed by an agent at a navigation step? 
(2) How can we reduce temporal redundancy 
across the agent's navigation steps?
(3) How can we use the mechanisms 
designed to address the two challenges 
to adaptively set early-exit thresholds of an existing method?
We present three novel techniques to address them individually.
In our evaluation, we demonstrate 
a 2--7.5$\times$ reduction in computations
while preserving performance across seven VLN benchmarks.
Moreover, we assess the robustness of our approach
under various visual corruptions that may occur in practice,
and identify challenges to address for future work.
We hope this work inspires future research on
developing efficient (and robust) VLN algorithms 
and promote their widespread adoption in real-world settings.

\section*{Acknowledgment}
\label{sec:ack}

We thank the anonymous reviewers for their valuable feedback.
This work is partially supported by the Samsung Global Research Outreach 2024 program.
The findings and conclusions in this work are those of the author(s) 
and do not necessarily represent the views of the funding agency.

{
    \small
    \bibliographystyle{ieeenat_fullname}
    \bibliography{bib/this_work}

\begin{thebibliography}{68}
\providecommand{\natexlab}[1]{#1}
\providecommand{\url}[1]{\texttt{#1}}
\expandafter\ifx\csname urlstyle\endcsname\relax
  \providecommand{\doi}[1]{doi: #1}\else
  \providecommand{\doi}{doi: \begingroup \urlstyle{rm}\Url}\fi

\bibitem[Anderson et~al.(2018)Anderson, Wu, Teney, Bruce, Johnson, Sünderhauf,
  Reid, Gould, and van~den Hengel]{R2R}
Peter Anderson, Qi Wu, Damien Teney, Jake Bruce, Mark Johnson, Niko
  Sünderhauf, Ian Reid, Stephen Gould, and Anton van~den Hengel.
\newblock Vision-and-language navigation: Interpreting visually-grounded
  navigation instructions in real environments.
\newblock In \emph{Proceedings of the IEEE Conference on Computer Vision and
  Pattern Recognition (CVPR)}, 2018.

\bibitem[Anderson et~al.(2021)Anderson, Shrivastava, Truong, Majumdar, Parikh,
  Batra, and Lee]{anderson2021simtoreal}
Peter Anderson, Ayush Shrivastava, Joanne Truong, Arjun Majumdar, Devi Parikh,
  Dhruv Batra, and Stefan Lee.
\newblock Sim-to-real transfer for vision-and-language navigation.
\newblock In \emph{Conference on Robot Learning}, pages 671--681. PMLR, 2021.

\bibitem[Andoni and Indyk(2008)]{andoni2008near}
Alexandr Andoni and Piotr Indyk.
\newblock Near-optimal hashing algorithms for approximate nearest neighbor in
  high dimensions.
\newblock \emph{Communications of the ACM}, 51\penalty0 (1):\penalty0 117--122,
  2008.

\bibitem[Banner et~al.(2019)Banner, Nahshan, and Soudry]{banner2019post}
Ron Banner, Yury Nahshan, and Daniel Soudry.
\newblock Post training 4-bit quantization of convolutional networks for
  rapid-deployment.
\newblock \emph{Advances in Neural Information Processing Systems}, 32, 2019.

\bibitem[Bay et~al.(2006)Bay, Tuytelaars, and Van~Gool]{bay2006surf}
Herbert Bay, Tinne Tuytelaars, and Luc Van~Gool.
\newblock Surf: Speeded up robust features.
\newblock In \emph{Computer Vision--ECCV 2006: 9th European Conference on
  Computer Vision, Graz, Austria, May 7-13, 2006. Proceedings, Part I 9}, pages
  404--417. Springer, 2006.

\bibitem[Bhalgat et~al.(2020)Bhalgat, Lee, Nagel, Blankevoort, and
  Kwak]{bhalgat2020lsq+}
Yash Bhalgat, Jinwon Lee, Markus Nagel, Tijmen Blankevoort, and Nojun Kwak.
\newblock Lsq+: Improving low-bit quantization through learnable offsets and
  better initialization.
\newblock In \emph{Proceedings of the IEEE/CVF conference on computer vision
  and pattern recognition workshops}, pages 696--697, 2020.

\bibitem[Chang et~al.(2017)Chang, Dai, Funkhouser, Halber, Niessner, Savva,
  Song, Zeng, and Zhang]{Matterport3D}
Angel Chang, Angela Dai, Thomas Funkhouser, Maciej Halber, Matthias Niessner,
  Manolis Savva, Shuran Song, Andy Zeng, and Yinda Zhang.
\newblock Matterport3d: Learning from rgb-d data in indoor environments.
\newblock \emph{International Conference on 3D Vision (3DV)}, 2017.

\bibitem[Charikar(2002)]{charikar2002similarity}
Moses~S Charikar.
\newblock Similarity estimation techniques from rounding algorithms.
\newblock In \emph{Proceedings of the thiry-fourth annual ACM symposium on
  Theory of computing}, pages 380--388, 2002.

\bibitem[Chattopadhyay et~al.(2021)Chattopadhyay, Hoffman, Mottaghi, and
  Kembhavi]{chattopadhyay2021robustnav}
Prithvijit Chattopadhyay, Judy Hoffman, Roozbeh Mottaghi, and Aniruddha
  Kembhavi.
\newblock Robustnav: Towards benchmarking robustness in embodied navigation.
\newblock In \emph{Proceedings of the IEEE/CVF International Conference on
  Computer Vision}, pages 15691--15700, 2021.

\bibitem[Chen et~al.(2021)Chen, Guhur, Schmid, and Laptev]{HAMT}
Shizhe Chen, Pierre-Louis Guhur, Cordelia Schmid, and Ivan Laptev.
\newblock History aware multimodal transformer for vision-and-language
  navigation.
\newblock \emph{Advances in neural information processing systems},
  34:\penalty0 5834--5847, 2021.

\bibitem[Chen et~al.(2022)Chen, Guhur, Tapaswi, Schmid, and
  Laptev]{chen2022think}
Shizhe Chen, Pierre-Louis Guhur, Makarand Tapaswi, Cordelia Schmid, and Ivan
  Laptev.
\newblock Think global, act local: Dual-scale graph transformer for
  vision-and-language navigation.
\newblock In \emph{Proceedings of the IEEE/CVF Conference on Computer Vision
  and Pattern Recognition}, pages 16537--16547, 2022.

\bibitem[Choi et~al.(2018)Choi, Wang, Venkataramani, Chuang, Srinivasan, and
  Gopalakrishnan]{choi2018pact}
Jungwook Choi, Zhuo Wang, Swagath Venkataramani, Pierce I-Jen Chuang,
  Vijayalakshmi Srinivasan, and Kailash Gopalakrishnan.
\newblock Pact: Parameterized clipping activation for quantized neural
  networks.
\newblock \emph{arXiv preprint arXiv:1805.06085}, 2018.

\bibitem[Choukroun et~al.(2019)Choukroun, Kravchik, Yang, and
  Kisilev]{choukroun2019low}
Yoni Choukroun, Eli Kravchik, Fan Yang, and Pavel Kisilev.
\newblock Low-bit quantization of neural networks for efficient inference.
\newblock In \emph{2019 IEEE/CVF International Conference on Computer Vision
  Workshop (ICCVW)}, pages 3009--3018. IEEE, 2019.

\bibitem[Cuturi(2013)]{cuturi2013sinkhorn}
Marco Cuturi.
\newblock Sinkhorn distances: Lightspeed computation of optimal transport.
\newblock \emph{Advances in neural information processing systems}, 26, 2013.

\bibitem[Devlin et~al.(2019)Devlin, Chang, Lee, and
  Toutanova]{devlin-etal-2019-bert}
Jacob Devlin, Ming-Wei Chang, Kenton Lee, and Kristina Toutanova.
\newblock {BERT}: Pre-training of deep bidirectional transformers for language
  understanding.
\newblock In \emph{Proceedings of the 2019 Conference of the North {A}merican
  Chapter of the Association for Computational Linguistics: Human Language
  Technologies, Volume 1 (Long and Short Papers)}, pages 4171--4186,
  Minneapolis, Minnesota, 2019. Association for Computational Linguistics.

\bibitem[Dosovitskiy et~al.(2021)Dosovitskiy, Beyer, Kolesnikov, Weissenborn,
  Zhai, Unterthiner, Dehghani, Minderer, Heigold, Gelly, Uszkoreit, and
  Houlsby]{ViT}
Alexey Dosovitskiy, Lucas Beyer, Alexander Kolesnikov, Dirk Weissenborn,
  Xiaohua Zhai, Thomas Unterthiner, Mostafa Dehghani, Matthias Minderer, Georg
  Heigold, Sylvain Gelly, Jakob Uszkoreit, and Neil Houlsby.
\newblock An image is worth 16x16 words: Transformers for image recognition at
  scale.
\newblock In \emph{International Conference on Learning Representations}, 2021.

\bibitem[Fan et~al.(2019)Fan, Grave, and Joulin]{fan2019reducing}
Angela Fan, Edouard Grave, and Armand Joulin.
\newblock Reducing transformer depth on demand with structured dropout.
\newblock \emph{arXiv preprint arXiv:1909.11556}, 2019.

\bibitem[Fang et~al.(2023)Fang, Ma, Song, Mi, and Wang]{fang2023depgraph}
Gongfan Fang, Xinyin Ma, Mingli Song, Michael~Bi Mi, and Xinchao Wang.
\newblock Depgraph: Towards any structural pruning.
\newblock In \emph{Proceedings of the IEEE/CVF conference on computer vision
  and pattern recognition}, pages 16091--16101, 2023.

\bibitem[Figurnov et~al.(2017)Figurnov, Collins, Zhu, Zhang, Huang, Vetrov, and
  Salakhutdinov]{figurnov2017spatially}
Michael Figurnov, Maxwell~D Collins, Yukun Zhu, Li Zhang, Jonathan Huang,
  Dmitry Vetrov, and Ruslan Salakhutdinov.
\newblock Spatially adaptive computation time for residual networks.
\newblock In \emph{Proceedings of the IEEE conference on computer vision and
  pattern recognition}, pages 1039--1048, 2017.

\bibitem[Fried et~al.(2018)Fried, Hu, Cirik, Rohrbach, Andreas, Morency,
  Berg-Kirkpatrick, Saenko, Klein, and Darrell]{fried2018speaker}
Daniel Fried, Ronghang Hu, Volkan Cirik, Anna Rohrbach, Jacob Andreas,
  Louis-Philippe Morency, Taylor Berg-Kirkpatrick, Kate Saenko, Dan Klein, and
  Trevor Darrell.
\newblock Speaker-follower models for vision-and-language navigation.
\newblock \emph{Advances in neural information processing systems}, 31, 2018.

\bibitem[Guhur et~al.(2021)Guhur, Tapaswi, Chen, Laptev, and
  Schmid]{guhur2021airbert}
Pierre-Louis Guhur, Makarand Tapaswi, Shizhe Chen, Ivan Laptev, and Cordelia
  Schmid.
\newblock Airbert: In-domain pretraining for vision-and-language navigation.
\newblock In \emph{Proceedings of the IEEE/CVF International Conference on
  Computer Vision}, pages 1634--1643, 2021.

\bibitem[Guo et~al.(2023)Guo, Stutz, and Schiele]{guo2023improving}
Yong Guo, David Stutz, and Bernt Schiele.
\newblock Improving robustness of vision transformers by reducing sensitivity
  to patch corruptions.
\newblock In \emph{Proceedings of the IEEE/CVF Conference on Computer Vision
  and Pattern Recognition}, pages 4108--4118, 2023.

\bibitem[Han et~al.(2015{\natexlab{a}})Han, Mao, and Dally]{han2015deep}
Song Han, Huizi Mao, and William~J Dally.
\newblock Deep compression: Compressing deep neural networks with pruning,
  trained quantization and huffman coding.
\newblock \emph{arXiv preprint arXiv:1510.00149}, 2015{\natexlab{a}}.

\bibitem[Han et~al.(2015{\natexlab{b}})Han, Pool, Tran, and
  Dally]{han2015learning}
Song Han, Jeff Pool, John Tran, and William Dally.
\newblock Learning both weights and connections for efficient neural network.
\newblock \emph{Advances in neural information processing systems}, 28,
  2015{\natexlab{b}}.

\bibitem[Hao et~al.(2020)Hao, Li, Li, Carin, and Gao]{hao2020towards}
Weituo Hao, Chunyuan Li, Xiujun Li, Lawrence Carin, and Jianfeng Gao.
\newblock Towards learning a generic agent for vision-and-language navigation
  via pre-training.
\newblock In \emph{Proceedings of the IEEE/CVF conference on computer vision
  and pattern recognition}, pages 13137--13146, 2020.

\bibitem[He et~al.(2016)He, Zhang, Ren, and Sun]{he2016deep}
Kaiming He, Xiangyu Zhang, Shaoqing Ren, and Jian Sun.
\newblock Deep residual learning for image recognition.
\newblock In \emph{Proceedings of the IEEE conference on computer vision and
  pattern recognition}, pages 770--778, 2016.

\bibitem[Hendrycks and Dietterich(2019)]{hendrycks2018benchmarking}
Dan Hendrycks and Thomas Dietterich.
\newblock Benchmarking neural network robustness to common corruptions and
  perturbations.
\newblock In \emph{International Conference on Learning Representations}, 2019.

\bibitem[Hoang and Liu(2023)]{hoang2023revisiting}
Duc~NM Hoang and Shiwei Liu.
\newblock Revisiting pruning at initialization through the lens of ramanujan
  graph.
\newblock \emph{ICLR 2023}, 2023.

\bibitem[Hong et~al.(2021)Hong, Wu, Qi, Rodriguez-Opazo, and Gould]{vlnbert}
Yicong Hong, Qi Wu, Yuankai Qi, Cristian Rodriguez-Opazo, and Stephen Gould.
\newblock A recurrent vision-and-language bert for navigation.
\newblock In \emph{Proceedings of the IEEE/CVF Conference on Computer Vision
  and Pattern Recognition (CVPR)}, pages 1643--1653, 2021.

\bibitem[Hong et~al.(2022)Hong, Wang, Wu, and Gould]{hong2022bridging}
Yicong Hong, Zun Wang, Qi Wu, and Stephen Gould.
\newblock Bridging the gap between learning in discrete and continuous
  environments for vision-and-language navigation.
\newblock In \emph{Proceedings of the IEEE/CVF conference on computer vision
  and pattern recognition}, pages 15439--15449, 2022.

\bibitem[Huang et~al.(2018)Huang, Chen, Li, Wu, van~der Maaten, and
  Weinberger]{msdnet}
Gao Huang, Danlu Chen, Tianhong Li, Felix Wu, Laurens van~der Maaten, and
  Kilian Weinberger.
\newblock Multi-scale dense networks for resource efficient image
  classification.
\newblock In \emph{International Conference on Learning Representations}, 2018.

\bibitem[Jacob et~al.(2018)Jacob, Kligys, Chen, Zhu, Tang, Howard, Adam, and
  Kalenichenko]{jacob2018quantization}
Benoit Jacob, Skirmantas Kligys, Bo Chen, Menglong Zhu, Matthew Tang, Andrew
  Howard, Hartwig Adam, and Dmitry Kalenichenko.
\newblock Quantization and training of neural networks for efficient
  integer-arithmetic-only inference.
\newblock In \emph{Proceedings of the IEEE conference on computer vision and
  pattern recognition}, pages 2704--2713, 2018.

\bibitem[Kamath et~al.(2023)Kamath, Anderson, Wang, Koh, Ku, Waters, Yang,
  Baldridge, and Parekh]{kamath2023newpath}
Aishwarya Kamath, Peter Anderson, Su Wang, Jing~Yu Koh, Alexander Ku, Austin
  Waters, Yinfei Yang, Jason Baldridge, and Zarana Parekh.
\newblock A new path: Scaling vision-and-language navigation with synthetic
  instructions and imitation learning, 2023.

\bibitem[Kaya et~al.(2019)Kaya, Hong, and Dumitras]{kaya2019shallow}
Yigitcan Kaya, Sanghyun Hong, and Tudor Dumitras.
\newblock Shallow-deep networks: Understanding and mitigating network
  overthinking.
\newblock In \emph{International conference on machine learning}, pages
  3301--3310. PMLR, 2019.

\bibitem[Krantz and Lee(2022)]{krantz2022sim}
Jacob Krantz and Stefan Lee.
\newblock Sim-2-sim transfer for vision-and-language navigation in continuous
  environments.
\newblock In \emph{European Conference on Computer Vision}, pages 588--603.
  Springer, 2022.

\bibitem[Krantz et~al.(2020)Krantz, Wijmans, Majundar, Batra, and
  Lee]{krantz2020vlnce}
Jacob Krantz, Erik Wijmans, Arjun Majundar, Dhruv Batra, and Stefan Lee.
\newblock Beyond the nav-graph: Vision and language navigation in continuous
  environments.
\newblock In \emph{European Conference on Computer Vision (ECCV)}, 2020.

\bibitem[Li and Bansal(2023)]{li2023improving}
Jialu Li and Mohit Bansal.
\newblock Improving vision-and-language navigation by generating future-view
  image semantics.
\newblock In \emph{Proceedings of the IEEE/CVF conference on computer vision
  and pattern recognition}, pages 10803--10812, 2023.

\bibitem[Li et~al.(2021)Li, Gong, Tan, Yang, Hu, Zhang, Yu, Wang, and
  Gu]{li2021brecq}
Yuhang Li, Ruihao Gong, Xu Tan, Yang Yang, Peng Hu, Qi Zhang, Fengwei Yu, Wei
  Wang, and Shi Gu.
\newblock Brecq: Pushing the limit of post-training quantization by block
  reconstruction.
\newblock \emph{arXiv preprint arXiv:2102.05426}, 2021.

\bibitem[Liu et~al.(2020)Liu, Zhou, Zhao, Wang, Deng, and Ju]{liu2020fastbert}
Weijie Liu, Peng Zhou, Zhe Zhao, Zhiruo Wang, Haotang Deng, and Qi Ju.
\newblock Fastbert: a self-distilling bert with adaptive inference time.
\newblock \emph{arXiv preprint arXiv:2004.02178}, 2020.

\bibitem[Louizos et~al.(2018)Louizos, Reisser, Blankevoort, Gavves, and
  Welling]{louizos2018relaxed}
Christos Louizos, Matthias Reisser, Tijmen Blankevoort, Efstratios Gavves, and
  Max Welling.
\newblock Relaxed quantization for discretized neural networks.
\newblock \emph{arXiv preprint arXiv:1810.01875}, 2018.

\bibitem[Lowe(2004)]{lowe2004distinctive}
David~G Lowe.
\newblock Distinctive image features from scale-invariant keypoints.
\newblock \emph{International journal of computer vision}, 60:\penalty0
  91--110, 2004.

\bibitem[Molchanov et~al.(2016)Molchanov, Tyree, Karras, Aila, and
  Kautz]{molchanov2016pruning}
Pavlo Molchanov, Stephen Tyree, Tero Karras, Timo Aila, and Jan Kautz.
\newblock Pruning convolutional neural networks for resource efficient
  inference.
\newblock \emph{arXiv preprint arXiv:1611.06440}, 2016.

\bibitem[Moudgil et~al.(2021)Moudgil, Majumdar, Agrawal, Lee, and
  Batra]{moudgil2021soat}
Abhinav Moudgil, Arjun Majumdar, Harsh Agrawal, Stefan Lee, and Dhruv Batra.
\newblock Soat: A scene-and object-aware transformer for vision-and-language
  navigation.
\newblock \emph{Advances in Neural Information Processing Systems},
  34:\penalty0 7357--7367, 2021.

\bibitem[Nagel et~al.(2020)Nagel, Amjad, Van~Baalen, Louizos, and
  Blankevoort]{nagel2020up}
Markus Nagel, Rana~Ali Amjad, Mart Van~Baalen, Christos Louizos, and Tijmen
  Blankevoort.
\newblock Up or down? adaptive rounding for post-training quantization.
\newblock In \emph{International Conference on Machine Learning}, pages
  7197--7206. PMLR, 2020.

\bibitem[Nova et~al.(2023)Nova, Dai, and Schuurmans]{nova2023gradient}
Azade Nova, Hanjun Dai, and Dale Schuurmans.
\newblock Gradient-free structured pruning with unlabeled data.
\newblock In \emph{International Conference on Machine Learning}, pages
  26326--26341. PMLR, 2023.

\bibitem[Perincherry et~al.(2025)Perincherry, Krantz, and
  Lee]{perincherry2025visual}
Akhil Perincherry, Jacob Krantz, and Stefan Lee.
\newblock Do visual imaginations improve vision-and-language navigation agents?
\newblock In \emph{Proceedings of the Computer Vision and Pattern Recognition
  Conference}, pages 3846--3855, 2025.

\bibitem[Qi et~al.(2020)Qi, Wu, Anderson, Wang, Wang, Shen, and
  Hengel]{qi2020reverie}
Yuankai Qi, Qi Wu, Peter Anderson, Xin Wang, William~Yang Wang, Chunhua Shen,
  and Anton van~den Hengel.
\newblock Reverie: Remote embodied visual referring expression in real indoor
  environments.
\newblock In \emph{Proceedings of the IEEE/CVF Conference on Computer Vision
  and Pattern Recognition}, pages 9982--9991, 2020.

\bibitem[Rublee et~al.(2011)Rublee, Rabaud, Konolige, and
  Bradski]{rublee2011orb}
Ethan Rublee, Vincent Rabaud, Kurt Konolige, and Gary Bradski.
\newblock Orb: An efficient alternative to sift or surf.
\newblock In \emph{2011 International conference on computer vision}, pages
  2564--2571. Ieee, 2011.

\bibitem[Savva et~al.(2019)Savva, Kadian, Maksymets, Zhao, Wijmans, Jain,
  Straub, Liu, Koltun, Malik, et~al.]{savva2019habitat}
Manolis Savva, Abhishek Kadian, Oleksandr Maksymets, Yili Zhao, Erik Wijmans,
  Bhavana Jain, Julian Straub, Jia Liu, Vladlen Koltun, Jitendra Malik, et~al.
\newblock Habitat: A platform for embodied ai research.
\newblock In \emph{Proceedings of the IEEE/CVF international conference on
  computer vision}, pages 9339--9347, 2019.

\bibitem[Sun et~al.(2024)Sun, Wang, Li, Cao, Jiang, Hu, and Zhang]{sun2024p4q}
Huixin Sun, Runqi Wang, Yanjing Li, Xianbin Cao, Xiaolong Jiang, Yao Hu, and
  Baochang Zhang.
\newblock P4q: Learning to prompt for quantization in visual-language models.
\newblock \emph{arXiv preprint arXiv:2409.17634}, 2024.

\bibitem[Tang et~al.(2023)Tang, Wang, Kong, Zhang, Li, Ding, Wang, Liang, and
  Xu]{MuE}
S. Tang, Y. Wang, Z. Kong, T. Zhang, Y. Li, C. Ding, Y. Wang, Y. Liang, and D.
  Xu.
\newblock You need multiple exiting: Dynamic early exiting for accelerating
  unified vision language model.
\newblock In \emph{2023 IEEE/CVF Conference on Computer Vision and Pattern
  Recognition (CVPR)}, pages 10781--10791, Los Alamitos, CA, USA, 2023. IEEE
  Computer Society.

\bibitem[Teerapittayanon et~al.(2016)Teerapittayanon, McDanel, and
  Kung]{teerapittayanon2016branchynet}
Surat Teerapittayanon, Bradley McDanel, and Hsiang-Tsung Kung.
\newblock Branchynet: Fast inference via early exiting from deep neural
  networks.
\newblock In \emph{2016 23rd international conference on pattern recognition
  (ICPR)}, pages 2464--2469. IEEE, 2016.

\bibitem[Thomason et~al.(2020)Thomason, Murray, Cakmak, and
  Zettlemoyer]{thomason2020vision}
Jesse Thomason, Michael Murray, Maya Cakmak, and Luke Zettlemoyer.
\newblock Vision-and-dialog navigation.
\newblock In \emph{Conference on Robot Learning}, pages 394--406. PMLR, 2020.

\bibitem[Uhlich et~al.(2019)Uhlich, Mauch, Cardinaux, Yoshiyama, Garcia,
  Tiedemann, Kemp, and Nakamura]{uhlich2019mixed}
Stefan Uhlich, Lukas Mauch, Fabien Cardinaux, Kazuki Yoshiyama, Javier~Alonso
  Garcia, Stephen Tiedemann, Thomas Kemp, and Akira Nakamura.
\newblock Mixed precision dnns: All you need is a good parametrization.
\newblock \emph{arXiv preprint arXiv:1905.11452}, 2019.

\bibitem[Wang et~al.(2022)Wang, Zhou, Zeng, and Zhang]{wang2022efficientvlm}
Tiannan Wang, Wangchunshu Zhou, Yan Zeng, and Xinsong Zhang.
\newblock Efficientvlm: Fast and accurate vision-language models via knowledge
  distillation and modal-adaptive pruning.
\newblock \emph{arXiv preprint arXiv:2210.07795}, 2022.

\bibitem[Wang et~al.(2018)Wang, Yu, Dou, Darrell, and
  Gonzalez]{wang2018skipnet}
Xin Wang, Fisher Yu, Zi-Yi Dou, Trevor Darrell, and Joseph~E Gonzalez.
\newblock Skipnet: Learning dynamic routing in convolutional networks.
\newblock In \emph{Proceedings of the European conference on computer vision
  (ECCV)}, pages 409--424, 2018.

\bibitem[Wang et~al.(2004)Wang, Bovik, Sheikh, and Simoncelli]{wang2004image}
Zhou Wang, Alan~C Bovik, Hamid~R Sheikh, and Eero~P Simoncelli.
\newblock Image quality assessment: from error visibility to structural
  similarity.
\newblock \emph{IEEE transactions on image processing}, 13\penalty0
  (4):\penalty0 600--612, 2004.

\bibitem[Wang et~al.(2023)Wang, Li, Hong, Wang, Wu, Bansal, Gould, Tan, and
  Qiao]{wang2023scaling}
Zun Wang, Jialu Li, Yicong Hong, Yi Wang, Qi Wu, Mohit Bansal, Stephen Gould,
  Hao Tan, and Yu Qiao.
\newblock Scaling data generation in vision-and-language navigation.
\newblock In \emph{Proceedings of the IEEE/CVF International Conference on
  Computer Vision}, pages 12009--12020, 2023.

\bibitem[Wang et~al.(2024)]{wang2024simtoreal}
Z. Wang et~al.
\newblock Sim-to-real transfer via 3d feature fields for vision-and-language
  navigation.
\newblock \emph{CoRL}, 2024.

\bibitem[Wasserman et~al.(2023)]{wasserman2023last}
J. Wasserman et~al.
\newblock Last-mile embodied visual navigation.
\newblock \emph{CoRL}, 2023.

\bibitem[Xin et~al.(2020)Xin, Tang, Lee, Yu, and Lin]{xin2020deebert}
Ji Xin, Raphael Tang, Jaejun Lee, Yaoliang Yu, and Jimmy Lin.
\newblock Deebert: Dynamic early exiting for accelerating bert inference.
\newblock \emph{arXiv preprint arXiv:2004.12993}, 2020.

\bibitem[Yue et~al.(2024)Yue, Wang, Kang, Han, Wang, Song, Feng, and
  Huang]{yue2024deer}
Yang Yue, Yulin Wang, Bingyi Kang, Yizeng Han, Shenzhi Wang, Shiji Song, Jiashi
  Feng, and Gao Huang.
\newblock Deer-vla: Dynamic inference of multimodal large language models for
  efficient robot execution.
\newblock \emph{Advances in Neural Information Processing Systems},
  37:\penalty0 56619--56643, 2024.

\bibitem[Zhang et~al.(2011)Zhang, Zhang, Mou, and Zhang]{zhang2011fsim}
Lin Zhang, Lei Zhang, Xuanqin Mou, and David Zhang.
\newblock Fsim: A feature similarity index for image quality assessment.
\newblock \emph{IEEE transactions on Image Processing}, 20\penalty0
  (8):\penalty0 2378--2386, 2011.

\bibitem[Zhang et~al.(2025)]{zhang2025humanoidpano}
Q. Zhang et~al.
\newblock Humanoidpano: Hybrid spherical panoramic-lidar cross-modal perception
  for humanoid robots.
\newblock \emph{arXiv}, 2025.

\bibitem[Zhang et~al.(2018)Zhang, Isola, Efros, Shechtman, and
  Wang]{zhang2018unreasonable}
Richard Zhang, Phillip Isola, Alexei~A Efros, Eli Shechtman, and Oliver Wang.
\newblock The unreasonable effectiveness of deep features as a perceptual
  metric.
\newblock In \emph{Proceedings of the IEEE conference on computer vision and
  pattern recognition}, pages 586--595, 2018.

\bibitem[Zhu et~al.(2021{\natexlab{a}})Zhu, Liang, Zhu, Yu, Chang, and
  Liang]{zhu2021soon}
Fengda Zhu, Xiwen Liang, Yi Zhu, Qizhi Yu, Xiaojun Chang, and Xiaodan Liang.
\newblock Soon: Scenario oriented object navigation with graph-based
  exploration.
\newblock In \emph{Proceedings of the IEEE/CVF Conference on Computer Vision
  and Pattern Recognition}, pages 12689--12699, 2021{\natexlab{a}}.

\bibitem[Zhu et~al.(2023)Zhu, An, Huang, and Hong]{pmlr-v202-zhu23a}
Sicheng Zhu, Bang An, Furong Huang, and Sanghyun Hong.
\newblock Learning unforeseen robustness from out-of-distribution data using
  equivariant domain translator.
\newblock In \emph{Proceedings of the 40th International Conference on Machine
  Learning}, pages 42915--42937. PMLR, 2023.

\bibitem[Zhu et~al.(2021{\natexlab{b}})]{zhu2021vigor}
S. Zhu et~al.
\newblock Vigor: Cross-view image geo-localization beyond one-to-one retrieval.
\newblock \emph{CVPR}, 2021{\natexlab{b}}.

\end{thebibliography}
}

\clearpage
\appendix
\section{Experimental Setup in Detail}
\label{appendix:exp-setup-detail}

We describe the experimental setup used to evaluate 
our input-adaptive inference mechanism in detail.
We implemented our strategy on top of the codebases provided by the authors of  HAMT~\cite{HAMT}, DUET~\cite{chen2022think}, and \vlncebert~\cite{krantz2022sim}.
During inference, instead of using cached image features, we integrate the original encoder (ViT-B/16~\cite{ViT} for HAMT and DUET and ResNet-152~\cite{he2016deep} for \vlncebert) to process the images directly.

\topic{Hardware and software.}
We run our experiments on a machine equipped with an Intel Xeon processor with 48 cores, 64GB of DRAM, and 8 NVIDIA A40 GPUs, with all inference tasks performed on a single GPU with a batch size of 1.
Following the original HAMT study, 
we use Python, PyTorch, and Cuda for all experiments, with versions in accordance with the original studies~\cite{chen2022think, HAMT, krantz2022sim}.
For GFLOPs calculations, we use the Python library thop\footnote{\url{https://pypi.org/project/thop}}.

\topic{Datasets.} We describe the benchmarks we use in detail:
\begin{itemize}[topsep=0.em, itemsep=0.1em, leftmargin=1.4em]
    \item \textbf{R2R}~\cite{R2R} is based on Matterport3D~\cite{Matterport3D}, 
    containing 10,567 panorama views taken from 90 photo-realistic houses.
    The dataset includes 7,189 shortest-path trajectories,
    each of which is associated with 3 natural language instructions.
    The training, validation (seen), validation (unseen), and test (unseen) sets
    include 61, 56, 11, and 18 houses, respectively.
    The validation (seen) set consists of houses in the training set, used to check the generalization status of a model during training, while the sets marked as `unseen' are the houses not in the training set.
    \item \textbf{R2R-Back}~\cite{HAMT} requires the agent to return to its starting point after reaching the destination. 
    To complete the task, the agent must remember its navigation history.
    A return command is appended to each R2R instruction, and the reversed path is provided as guidance for the return trip.
    \item \textbf{R2R-Last}~\cite{HAMT} uses only the last sentence from the original R2R instructions to describe the destination.
    
    \item \textbf{REVERIE}~\cite{qi2020reverie} provides high-level instructions, closer to those given by humans, replacing the step-by-step instructions of R2R.
    Instead of navigating to a target location, 
    the agent is required to identify and localize the target object upon arrival, making the task more complex and realistic.
    The dataset includes 4,140 target objects, which are categorized into 489 distinct groups.
    \item \textbf{CVDN}~\cite{thomason2020vision} requires the agent to navigate based on long, 
    potentially unclear instructions.
    The agent interacts with a navigator 
    through question and answer dialog 
    to clarify and complete the task.
    In total, it has 2,050 human-human navigation dialogues, 
    consisting of over 7,000 navigation trajectories 
    accompanied by question-answer interactions, 
    covering 83 matterport3D houses.
    \item \textbf{SOON}~\cite{zhu2021soon} is similar to REVERIE
    but contains longer and more detailed instructions.
    The average length of these instructions is 47 words, 
    with path lengths varying from 2 to 21 steps.
    It requires the agent to navigate by 
    understanding the relationship between objects 
    in the environment to accurately locate the target object.
    \item \textbf{R2R-CE}~\cite{krantz2020vlnce} is a continuous version of R2R supported by the Habitat simulator. 
    To generate the dataset, Krantz \textit{et al.} convert the static panoramic scene data in Matterport3D into a continuous environment using mesh-based 3D reconstructions. 
    R2R trajectories are then transferred by mapping their nodes to the closest navigable locations on the reconstructed mesh. 
    Non-navigable nodes (e.g., those placed on furniture or spanning disjoint regions) were filtered out. 
    The final dataset consists of 4475 successfully transferred trajectories, each paired with the original R2R instruction set. Note that unlike the original R2R setting, where agents teleport between nodes, R2R-CE requires agents to navigate using low-level actions such as moving forward and turning.
\end{itemize}

\section{Optimal Hyperparameters for Adapting MuE}
\label{appendix:mue-with-diff-thresholds}

\begin{figure}[ht]
\centering
\begin{minipage}[t]{0.54\linewidth}
    \centering
    \includegraphics[width=\linewidth]{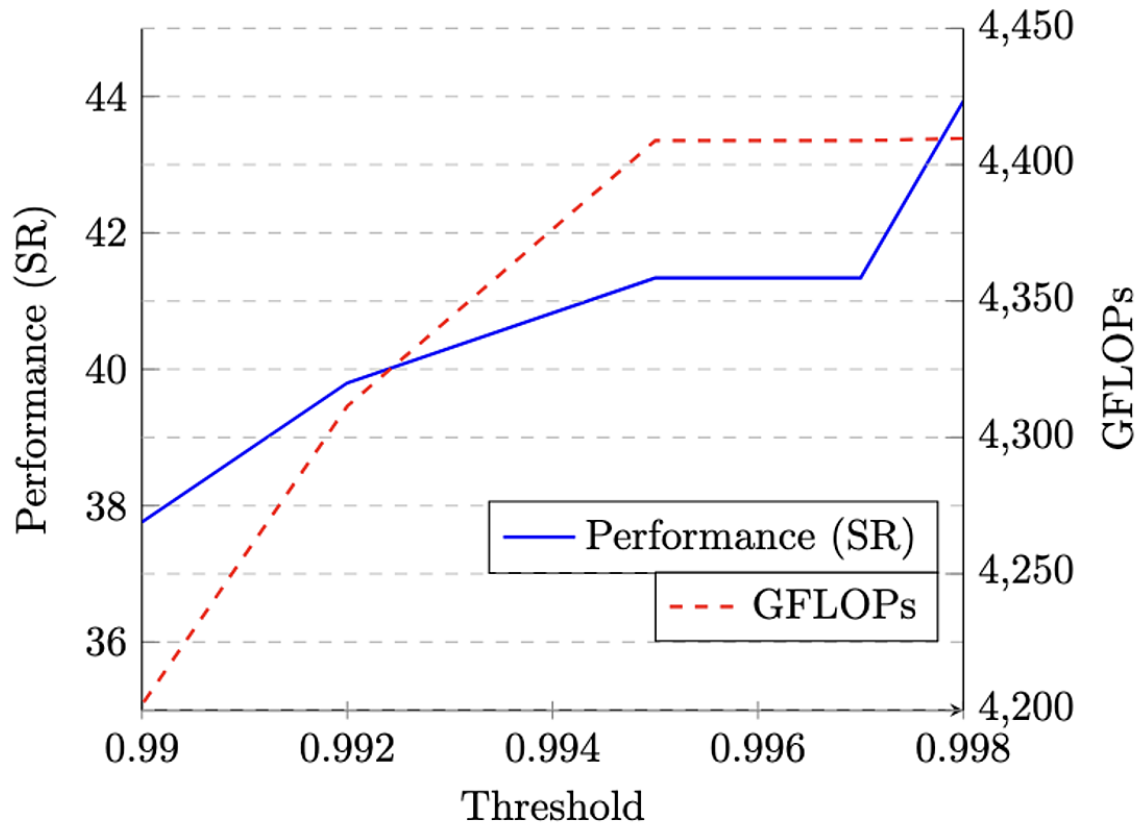}
    \caption{%
        \textbf{Comparison of performance (in SR) and GFLOPs in MuE 
        across different thresholds.}
    }
    \label{fig:MuE_thresholds}
\end{minipage}
\hfill
\begin{minipage}[t]{0.43\linewidth}
    \centering
    \includegraphics[width=\linewidth]{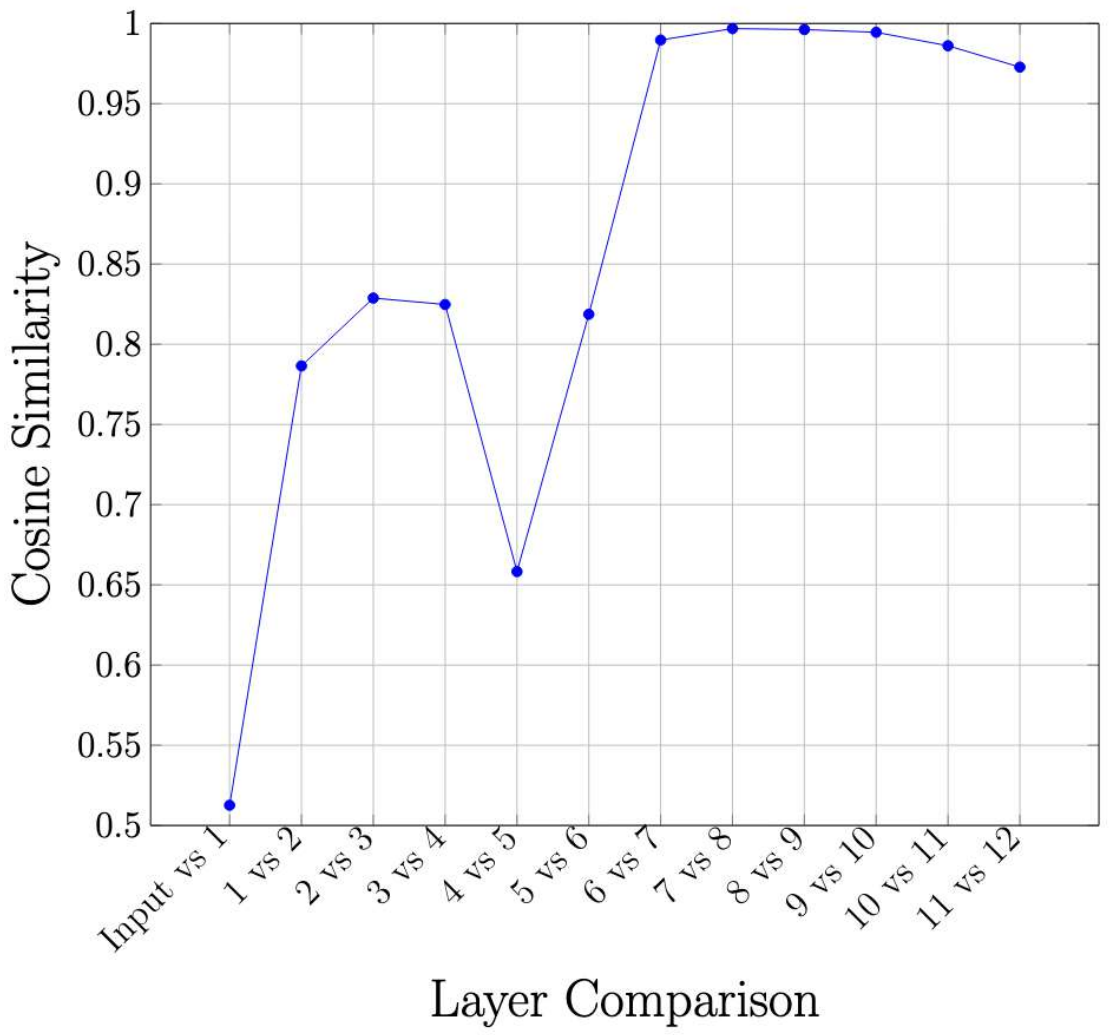}
    \caption{%
        \textbf{Cosine similarity between adjacent layers of ViT used in HAMT.}
    }
    \label{fig:ViT_saturation}
\end{minipage}
\end{figure}

To best evaluate MuE on VLN tasks, we perform a hyperparameter sweep over the early-exit threshold.
Figure~\ref{fig:MuE_thresholds} presents the performance (in SR) and GFLOPs across different early exit thresholds applied to 
the MuE version of ViT used in the HAMT agent, tested on the R2R dataset.
The lowest threshold we report is 0.99, 
as lower thresholds caused a dramatic drop in performance (more than 50\%).
As the threshold increases, 
the success rate of the MuE agent increases substantially but at the cost of computational savings.
Even for thresholds close to 1, meaning that the ViT is using a majority of its layers for each input, we still see a large performance drop compared to the baseline agent. As we discuss in Sec~\ref{subsec:our-method}, this is likely because MuE statically applies early-exits, causing it to under-process important components of the panorama such as navigable views.

\topic{Why does MuE underprocess important views?}
The intuition behind MuE~\cite{MuE} is that the
activations of Transformer-based vision models \emph{saturate},
where their similarity between layers peaks early on,
and is maintained at future stages of computation, suggesting 
a lack of new/useful information.
MuE then exploits this property to skip
the later layers without a significant loss in performance.
So, for MuE to be successful, the similarity of activations must sufficiently saturate and not decrease at later layers.
However, as shown in Figure~\ref{fig:ViT_saturation},
the necessary saturation pattern is not observed in the VLN setting.
The cosine similarity peaks between layers 7 and 8 
but then decreases for all future layers.
This explains the significant performance drop when MuE is directly applied to VLN agents, as it consistently early-exits despite saturation not being achieved.

\begin{figure}[ht]
\begin{minipage}{\linewidth}
\vspace{-1.2em} 
\begin{algorithm}[H]
\caption{SimHash Algorithm}
\label{alg:simhash-algorithm}
\begin{algorithmic} [1]
\Statex\hspace{-1.5em}\textbf{Input:} 
    a current view $v_i$
\Statex\hspace{-1.5em}\textbf{Output:} 
    a binary hash $key$
\Function{Hash}{$v_i$}
    \State $key \gets \varnothing$
    \For{each $hp$ \textbf{in} Hyperplanes}
        \State $sign \gets \texttt{DotProduct}(hp, v_i)$
        \State $hash\_val \gets (sign > 0)$ \Comment{converts to binary}
        \State $key \gets key + hash\_val$
    \EndFor
    \State \Return $key$
\EndFunction

\Statex\hspace{-1.5em}\textbf{Input:} 
    a hash table $h$, 
    a current view $v_i$,
    an embedding $e_i$
\Statex\hspace{-1.5em}\textbf{Output:} 
a hash table $h$
\Function{AddToHashTable}{$h, v_i, e_i$}
    \State $key \gets \texttt{Hash}(v_i)$
    \State $h \gets \texttt{InsertToHashTable}(key, v_i, e_i)$
    \State \Return $h$
\EndFunction
\Statex\hspace{-1.5em}\textbf{Input:} 
    a hash table $h$, 
    a current view $v_i$
\Statex\hspace{-1.5em}\textbf{Output:} 
    an embedding $e_i$ 
\Function{FindSimilar}{$h, v_i$}
    \State $s_{max} \gets -1$
    \State $key \gets \texttt{Hash}(v_i)$
    \State $bucket \gets h.get(key)$
        \For{each $(v_{candidate}, e_{candidate})$ \textbf{in} bucket}
            \State $s \gets$ \textbf{CosineSimilarity}($v_i$, $v_{candidate}$)
            \If{$s > s_{max}$}
                \State $s_{max} \gets s$
                \State $e_{best} \gets e_{candidate}$
            \EndIf
        \EndFor
        \If{$s_{max} > threshold$}
            \State $e_i \gets e_{best}$
        \Else
            \State $e_i \gets \varnothing$
        \EndIf
        \State \Return $e_i$
\EndFunction
\end{algorithmic}
\end{algorithm}
\vspace{-2.0em}
\end{minipage}
\end{figure}

\section{Our LSH Algorithm in Detail}
\label{appendix:rplsh-algorith-detail}

A core mechanism we introduce in Sec~\ref{subsubsec:harnessing-temporal-locality} is our SimHash algorithm, used to avoid reprocessing previously seen and similar images. Algorithm~\ref{alg:simhash-algorithm} details our implementation.

\topic{(line 1-9) Hashing RGB vectors.} Given an image, we first hash the raw RGB vector into a short binary encoding using random projection~\cite{charikar2002similarity, andoni2008near}. The algorithm calculates the dot product between the image vector and each hyperplane. If the dot product is positive, it assigns a binary value of 1, otherwise it assigns 0. These binary values are sequentially appended to form a complete binary hash key. The length of the hash key is determined by the number of hyperplanes used in the projection.

\topic{(line 10-14) Adding embeddings to the hash table.} This function is used to insert processed images and their corresponding embeddings into the hash table for future use.

\topic{(line 15-32) Retrieving a similar embedding.} This function takes an image we have not yet processed and tries to find a suitable embedding candidate. We first obtain all embeddings with images similar to the current image by hashing it into its binary encoding and accessing the corresponding bucket in the hash table. We then loop through all images associated with the similar embeddings and find the one yielding the highest similarity score (in our main experiments, the score is computed using cosine similarity). If this score exceeds our threshold hyperparameter, we return the associated embedding; otherwise, we return nothing.

\topic{Running the algorithm.} We employ the above three functions to run SimHash on an arbitrary panorama. For each extended navigable view (other views are omitted and explained in Algorithm~\ref{alg:our-algorithm}), we attempt to use a high-similarity embedding from the hash table. If it exists, we reuse this embedding for the current view and continue to the next. If not, we need to process the view using the ViT adapted for MuE, and then add the image and its embedding to the hash table. After processing the entire panorama, we return the set of final embeddings to be used for agent navigation.

\topic{Storage overhead analysis.} Here, we consider the storage overhead necessary to deploy our hashing algorithm on VLN agents.
Our LSH technique stores pairs of images and embeddings. In the benchmarks we consider, these images are of size 3x224x224 (Matterport3D) or 3x480x640 (Habitat). 
The embedding size depends on the model: 197x768 for HAMT and DUET  (the number of ViT patches times the model’s hidden dimension) and 2048 for \vlncebert (the hidden dimension of ResNet-152). 
These are stored in full-precision floating-point format (4 bytes per value), resulting in $(3 \times 224 \times 224 + 197 \times 768) \times 4 \approx 1.2$ MB for HAMT and DUET and $(3 \times 480 \times 640 + 2048) \times 4 \approx 3.7$ MB for \vlncebert per cached pair. 
For standard VLN, the longest navigation route was $\sim$12 steps (from R2R-Back). Assuming caching of all 36 images per panorama, the worst-case storage overhead is 522.7 MB.
However, in practice, most tasks involve 5--7 steps, and we cache at most 14 images per step, yielding a more typical overhead of 84.7--118.6 MB.
For continuous VLN, the longest navigation route is $\sim$130 steps, and we cache at most 6 views, leading to a worst-case overhead of 2.9 GB. The average trajectory length is 56 steps, with about 3 views cached per step, resulting in an average overhead of 609.6 MB.
Given that modern DRAM sizes are orders of magnitude larger, this storage overhead remains manageable for practical deployment.

\section{Full Standard VLN Evaluation Results}
\label{appendix:full-results}

\begin{table}[h]
\centering
\adjustbox{max width=\linewidth}{
\begin{tabular}{@{}lclcccccc@{}}
\toprule
 \multirow{2}{*}{\textbf{Agent}} & \multirow{2}{*}{\textbf{Task}} & \multirow{2}{*}{\textbf{Method}} & \multicolumn{5}{c}{\textbf{Performance}} & \multirow{2}{*}{\textbf{\Ourmetric{}}} \\ \cmidrule(l){4-8} 
 &  &  & \textbf{TL} & \textbf{OSR} & \textbf{SR} & \textbf{SPL} & \textbf{GP} & \\ \midrule \midrule
 \multirow{9}{*}{HAMT}&
  \multirow{2}{*}{\textbf{R2R}}& Base & 11.53 & 74.29 & 66.16 & 61.49 & - & 4763.24 \\
 & & Ours (All)& 12.87 & 71.95 & 60.41 & 54.50 & - & 1917.61 \\ \cmidrule{3-9}& 
 \multirow{2}{*}{\textbf{R2R-Back}}& Base& 20.56& -& 55.43 &52.34&  -&8181.55\\
 & & Ours (All)& 20.53& -& 49.21 &46.47&  -&3331.80\\ \cmidrule{3-9}& 
 \multirow{2}{*}{\textbf{R2R-Last}}& Base& 12.28& 54.24& 47.85 &42.27& -&4982.68\\
 & & Ours (All)& 12.36& 49.72& 41.93 &36.97& -&2589.44\\ \cmidrule{3-9}& 
 \multirow{2}{*}{\textbf{CVDN}}& Base& -& -& - &-& 4.88&11022.03\\
 & & Ours (All)& -& -& - &-& 4.45&4773.34\\ \midrule \midrule
 \multirow{5}{*}{DUET}&\multirow{2}{*}{\textbf{R2R}}& Base & 13.94 & 81.10 & 71.73 & 60.57 & - & 4998.00 \\
 & & Ours (All)& 14.21 & 73.86 & 63.47 & 52.35 & - & 2026.30 \\ \cmidrule{3-9}& 
 \multirow{2}{*}{\textbf{SOON}}& Base& 35.87& 50.38& 36.19 &22.67&  -&9997.81+$C$\\
 & & Ours (All)& 42.36& 54.22&  36.43&20.37&  -&4533.83+$C$\\
 \bottomrule
\end{tabular}
}
\vspace{0.1em}
\caption{%
\textbf{Performance and efficiency of the baseline agents versus our improved-efficiency agents across multiple benchmarks.} We denote the cost of object feature extraction as $C$.}
\label{tbl:appendix-results}
\end{table}

Table~\ref{tbl:appendix-results} complements our main evaluation of standard VLN in Sec~\ref{subsec:effectiveness} with additional benchmarks: R2R~\cite{R2R}, R2R-Back~\cite{HAMT}, R2R-Last~\cite{HAMT}, CVDN~\cite{thomason2020vision}, and SOON~\cite{zhu2021soon}. 
For CVDN, we report the additional evaluation metric Goal Progress (GP), which 
assigns a higher score as the agent moves closer to the goal, indicating better performance~\cite{HAMT}.

The upper section of the table compares the performance and efficiency of the baseline and our efficient HAMT agents.
For R2R and R2R-Back, our strategy reduces computations by 60\% with an SR drop of 9--11\%.
For R2R-Last, we reduce computation by 48\%, with a 12\% reduction in SR.
Finally, for the CVDN evaluation, our efficient model reduces computation by 57\%, with only a 9\% decrease in GP.

The lower section of the table presents a comparison of the performance and efficiencies of the DUET agents. 
For R2R, our strategy achieves a 59\% speed-up with a 12\% decrease in SR.
For SOON, we observed a marginal increase in SR accompanied by a 10\% drop in SPL, while saving 5463.98 GFLOPs (a 55\% reduction in visual feature processing).
These results demonstrate that our efficiency strategies are applicable across different benchmarks, achieving substantial computational savings while maintaining an acceptable trade-off in performance.

\begin{table}
\adjustbox{max width=\linewidth}{
\begin{tabular}{@{}lcccc@{}}
\toprule
\textbf{Agent} & \textbf{Task} & \textbf{Average Path Length} & \textbf{$\Delta$NE($\downarrow$)} & \textbf{$\Delta$GFLOPs($\downarrow$)} \\ \midrule \midrule
\multirow{3}{*}{HAMT} & \textbf{R2R}       & 6.0  & +0.53 & -2845.63 \\ 
                      & \textbf{R2R-Last}  & 6.0  & +0.45 & -2393.24 \\ 
                      & \textbf{R2R-Back}  & 12.0 & +0.54 & -5463.98 \\ \midrule
\multirow{2}{*}{DUET} & \textbf{R2R}      & 6.0  & +0.68 & -2971.70 \\ 
                      & \textbf{SOON}      & 9.6  & -0.44 & -5463.98 \\ 
\bottomrule
\end{tabular}
}
\vspace{0.1em}
\caption{%
\textbf{Performance of our efficient HAMT agent on benchmarks with different path lengths.} $\Delta$NE and $\Delta$GFLOPs are the changes in navigation error (NE) and GFLOPs compared to the baseline agent. The path length is the minimum number of navigation actions needed to reach the target destination.}
\label{tbl:path-length-results}
\end{table}

\topic{Robustness to navigation length.} It is possible that the errors introduced by our method \emph{propagate}, resulting in worse agent navigation for longer trajectories. We study if this is the case by considering the \emph{navigation error} (NE)---the distance of an agent's final position to the target position (in meters)---on benchmarks with varying path lengths. We deploy all of our proposed methods (simultaneously) on the HAMT agent and report the changes in NE and GFLOPs compared to the baseline in Table~\ref{tbl:path-length-results}.

We find our method is largely robust to longer path lengths. The NE does not increase for longer trajectories, and we even see a decrease for the SOON benchmark, which has an average path length 3.6 more steps than R2R. The results also show that our efficient VLN agent sees roughly proportional computational savings for longer paths. For example, the average path length in R2R-Back is double R2R, and we achieve a 1.92x larger reduction in GFLOPs for the HAMT agent.

\begin{table}[ht]
\centering
\adjustbox{max width=\linewidth}{
\begin{tabular}{lclc@{}}  
\toprule
  \textbf{Task}&\textbf{Agent}&  \textbf{Method}& \textbf{Wall-time (s)}\\ 
 \midrule \midrule
  \multirow{4}{*}{\textbf{R2R}}& \multirow{2}{*}{HAMT} & Base & 200811\\
    & & Ours & 119514\\ 

  \cmidrule{2-4}& \multirow{2}{*}{DUET}& Base & 268962\\ 
  & & Ours & 170464\\
 \bottomrule
\end{tabular}
}
\vspace{0.5em}
\caption{%
\textbf{Wall-time comparison} between the baseline agent and our efficient agent on the R2R task.}
\label{tbl:wall-time_comparison}
\end{table}

\topic{Runtime comparison.}
To validate that our approach improves efficiency in the real world, we report the wall-time comparison between our efficient VLN model and the baseline VLN for both HAMT and DUET agents, tested on the R2R validation unseen split, in Table~\ref{tbl:wall-time_comparison}.
Evidently, our efficient strategy applied to the VLN agents results in significant runtime savings, with an approximate 40\% reduction.
It is important to note that the disparity between the 60\% GFLOPs savings and the 40\% runtime reduction can be attributed to various hardware and software-related factors, such as simulation overhead, memory bandwidth limitations, or cache latency.

\begin{table}[ht]
\centering
\adjustbox{max width=\linewidth}{
    \begin{tabular}{@{}lccccc@{}}
    \toprule
    \textbf{Method}              & \textbf{TL}$(\downarrow)$   & \textbf{OSR}$(\uparrow)$   & \textbf{SR}$(\uparrow)$    & \textbf{SPL}$(\uparrow)$   & \textbf{GFLOPs}$(\downarrow)$ \\ \midrule \midrule
    None (Base) & 11.53& 74.29& 66.16& 61.49& 4763.24 \\ \midrule 
    $k$-extension         & 12.52         & 71.86          & 61.30          & 55.79          & 2,408.99        \\
    thresholds          & 12.33         & 72.46          & 62.62          & 57.39          & 3,867.46        \\
    LSH                 & 11.53         & 74.20          & 66.11          & 61.47          & 3,894.76        \\
    $k$-extension+LSH     & 12.52         & 71.90          & 61.17          & 55.63          & 2,013.48        \\
    $k$-extension+thresholds & 12.89      & 71.95          & 60.41          & 54.57          & 2,294.23        \\
    thresholds+LSH      & 12.33         & 72.41          & 62.49          & 57.33          & 3,190.66        \\
    All                & 12.87         & 71.95          & 60.41          & 54.50          & 1,917.61        \\ \bottomrule
    \end{tabular}
}
\vspace{0.1em}
\caption{\textbf{Performance of all combinations of our speed-up techniques} ($k$-extensions, early-exiting, and LSH) with the HAMT agent on the R2R benchmark.}
\label{tbl:per-mechanism-results}
\end{table}

\section{Per-Mechanism Analysis}
\label{appendix:per-mech-analysis}

In most experiments, we treat our proposed mechanisms as a single unit by applying all three simultaneously. While this is the most flexible and offers the best trade-off between performance and efficiency, analyzing each mechanism independently can provide valuable insights into its concise impact. Here, we present results on a per-mechanism basis.

\topic{Effectiveness.}
In Sec~\ref{subsec:effectiveness}, we apply our $k$-extension technique and then add adaptive thresholding early-exiting (denoted thresholds in Table~\ref{tbl:main-results}) and locality-sensitive hashing (LSH) as we found those combinations of techniques offer the most computational savings. Here, we study all combinations of three efficiency mechanisms. To use early-exiting and LSH without $k$-extension, we treat every non-navigable view as one that can be early-exited or hashed. Navigable views are still fully processed. We report results for the HAMT agent on the R2R benchmark in Table~\ref{tbl:per-mechanism-results}.

The results show that between individual techniques, $k$-extension offers the best computational savings with a 49\% reduction compared to the baseline agent. Early-exiting and LSH only reduce GFLOPs by $\sim$18\% because early-exiting still requires processing every view, and LSH reuses only a minority of cached image embeddings. We find that LSH provides better performance than the other two individual mechanisms, with an SR only 0.05 lower than the baseline. This is likely because the cached embeddings reused by LSH are near-identical, having a negligible impact on performance when interchanged. However, it is far less efficient than when combined with our other techniques. 

The combination we do not present in Table~\ref{tbl:main-results}, early-exiting and LSH (\textbf{thresholds+LSH}), provides slightly better performance than combinations using $k$-extension but at the cost of 39--66\% more GFLOPs. This suggests that retaining and partially processing/reusing the non-navigable views mitigates performance drop but is not nearly as efficient as $k$-extension. Overall, we find that all combinations of our techniques fare well, offering different trade-offs between performance and efficiency.

\topic{Robustness to natural corruptions.}
Now, we complement Sec~\ref{subsec:robustness} and study the robustness of each of our proposed mechanisms to visual corruption. We select the Low Lighting and Motion Blur corruptions based on their varying impact on performance and being more likely to occur in real-world VLN systems. We apply our methods to the HAMT agent and report results on R2R in Table~\ref{tbl:per-mech-visual-corruption}.

Our methods appear more robust to Low Lighting than Motion Blur, which corroborates our findings in Sec~\ref{subsec:robustness}. Across both corruptions, $k$-extension and early-exiting see a slight increase of 150--200 GFLOPs compared to the results in Table~\ref{tbl:per-mechanism-results}. This can likely be attributed to the increased trajectory length, and for early-exiting, we also find that the OOD samples require more ViT layers before sufficiently saturating. Both mechanisms result in significant drops in performance, though less than when we apply all simultaneously (results shown in Table~\ref{tbl:OOD}). 
Early-exiting is slightly more robust, achieving a 2--7\% higher SR, which makes sense as it processes strictly more images than $k$-extension. 

Interestingly, LSH functions extremely well when Low Lighting is applied. It offers a $\sim$49\% reduction in GFLOPs, compared to just 18\% when no corruption is present. We hypothesize that the reduced lighting makes more images similar, causing our algorithm to find more matches and reuse more embeddings. It also offers significant robustness, only incurring a 1\% point drop in SR. It seems like our caching mechanism is better suited for this environment, a finding we hope to explore in future work. For Motion Blur, LSH is less successful, being more robust than our other mechanisms but with minimal computational savings.

\begin{table}[ht]
\centering
\adjustbox{max width=\linewidth}{
    \begin{tabular}{@{}clccccc@{}}
    \toprule
    \textbf{Corruption}                    & \textbf{Method}        & \textbf{TL}($\downarrow$)    & \textbf{OSR}($\uparrow$)   & \textbf{SR}($\uparrow$)    & \textbf{SPL}($\uparrow$)   & \textbf{GFLOPs}($\downarrow$)  \\ \midrule \midrule
    \multirow{4}{*}{Low Lighting} & None (Base) & 12.15 & 71.31 & 62.58 & 57.23 & 4903.06         \\ \cmidrule(l){2-7} 
                                  & $k$-extension & 13.86 & 71.14 & 57.34 & 50.78 & 2571.06 \\
                                  & thresholds    & 13.63 & 70.29 & 58.79 & 52.16 & 4099.21 \\
                                  & LSH           & 12.95 & 71.43 & 61.47 & 55.19 & 2444.05 \\ \midrule
    \multirow{4}{*}{Motion Blur}  & None (Base) & 12.41 & 68.20 & 59.13 & 54.01 & 4996.64         \\ \cmidrule(l){2-7} 
                                  & $k$-extension & 14.03 & 65.13 & 53.77 & 48.01 & 2588.06 \\
                                  & thresholds    & 13.81 & 68.20 & 57.51 & 51.05 & 4073.04 \\
                                  & LSH           & 12.39 & 68.03 & 59.30 & 54.04 & 4030.52 \\ \bottomrule
    \end{tabular}
}
\vspace{0.1em}
\caption{\textbf{Performance under visual corruption of our methods applied \emph{independently}} to the HAMT agent on the R2R benchmark.}
\label{tbl:per-mech-visual-corruption}
\end{table}

\section{Information Loss Analysis}

In this section, we explore what types of information are lost when applying each of our speed-up techniques.

\begin{figure}
    \centering
    \includegraphics[width=\linewidth]{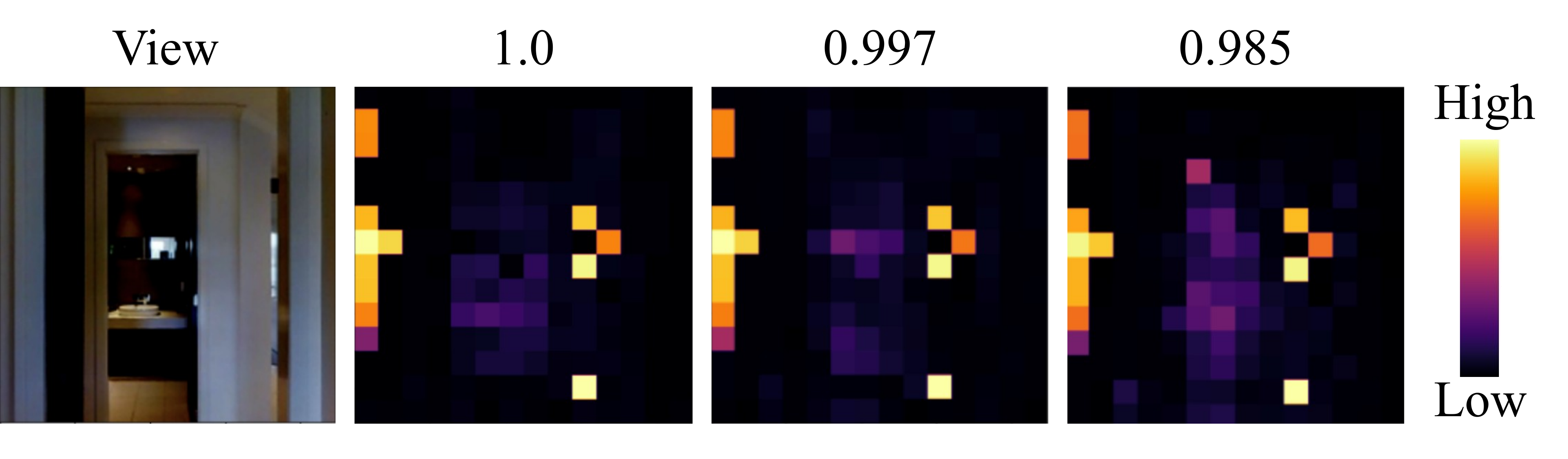}
    \vspace{-2em}
    \caption{\textbf{Attention visualization} across different exit thresholds on HAMT. Lower thresholds use fewer ViT layers.}
    \label{fig:attention-vis}
\end{figure}

\topic{Multi-exiting with thresholds.}
To assess the effect of processing views through fewer ViT layers, we analyze attention visualizations.
Figure~\ref{fig:attention-vis} illustrates attention maps from our efficient HAMT agent on a representative view.
In this example, the correct action is to ignore the bathroom and move to the side. 
As the exit layer decreases, HAMT focuses more on the bathroom, indicating a slight degradation in visual understanding.
However, this change is minimal, and the overall navigation outcome is unaffected.
Therefore, our adaptive thresholding technique provides an effective trade-off between computational efficiency and visual fidelity.

\topic{$k$-extensions.} 
For $k$-extensions, we fully mask non-navigable views, making local attention visualizations uninformative. 
To capture the \emph{global} impact of this masking, we measure the change in embeddings after processing through the cross-modal transformer.
Specifically, we extract visual, language, and history embeddings from 100 navigation steps of HAMT on R2R. 
We then apply the $k$-extensions technique, re-extract the embeddings, and compute the mean L2 distance for each navigation step. 
To ensure comparability, we only consider the first step in each environment, as subsequent steps may diverge.

\begin{figure}
    \centering
    \includegraphics[width=\linewidth]{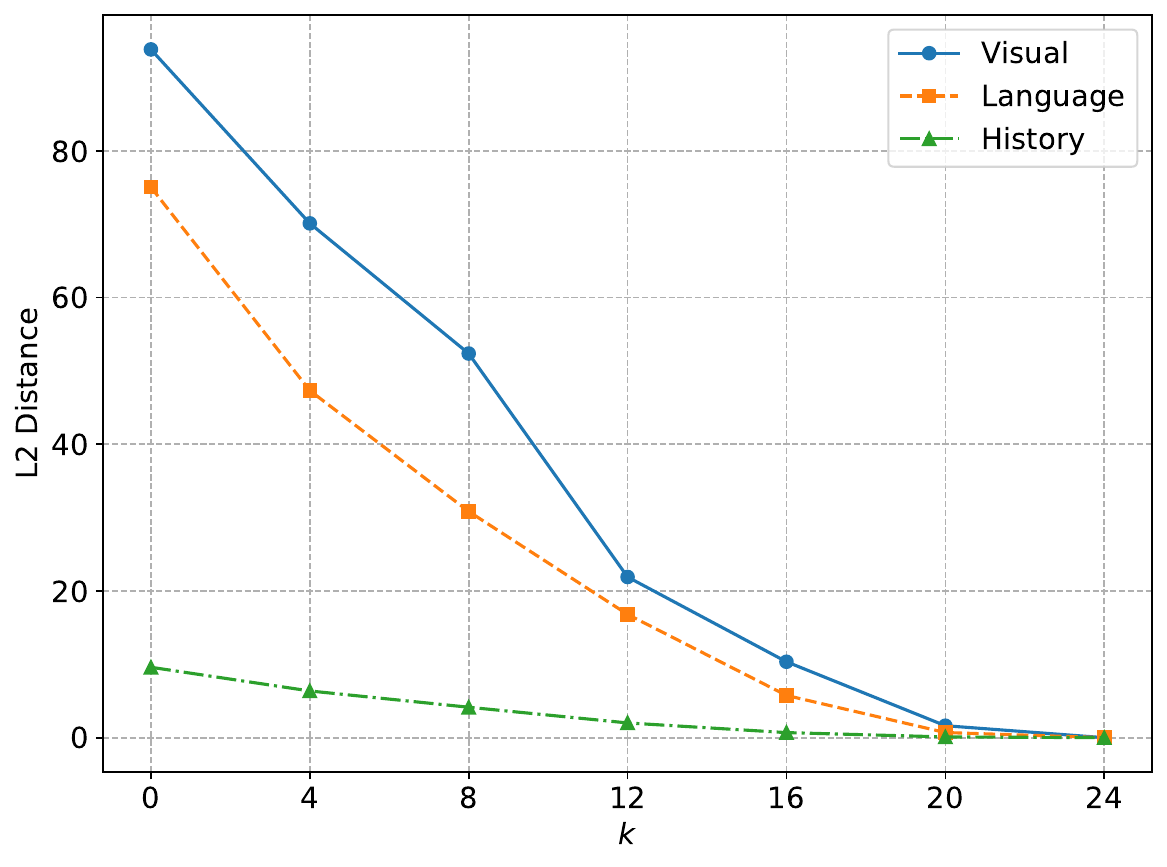}
    \vspace{-2em}
    \caption{\textbf{L2 distance of cross-modal embeddings} from the baseline and our efficient HAMT agent for different $k$ values. Embeddings are computed on 100 navigation instructions from R2R.}
    \label{fig:l2-distance-of-embeddings}
\end{figure}

Figure~\ref{fig:l2-distance-of-embeddings} shows the results across different values of $k$. 
As $k$ increases (i.e., more views are processed), the L2 distance for all embedding types decreases significantly. 
Interestingly, while $k=4$--$6$ only marginally reduces these distances, we still observe strong performance in Sec.~\ref{subsec:effectiveness}. 
This suggests that much of the information captured by these embeddings is not critical for navigation—an insight we leverage for computational efficiency.
Masking views affects visual embeddings the most, as they consistently have the highest L2 distance for all values of $k$.
However, we observe that language embeddings, which encode the navigation instructions, are also notably impacted. 
This further explains the performance degradation: if the agent does not understand the instruction, it may fail to navigate or stop appropriately.
In contrast, history embeddings are more resilient, likely because we only evaluate the first navigation step where historical context is minimal.
Overall, these results indicate that masking views leads to information loss that extends beyond visual perception. However, this loss is not critical for effective navigation with the appropriate choice of $k$.

\begin{figure}
    \centering
    \includegraphics[width=\linewidth]{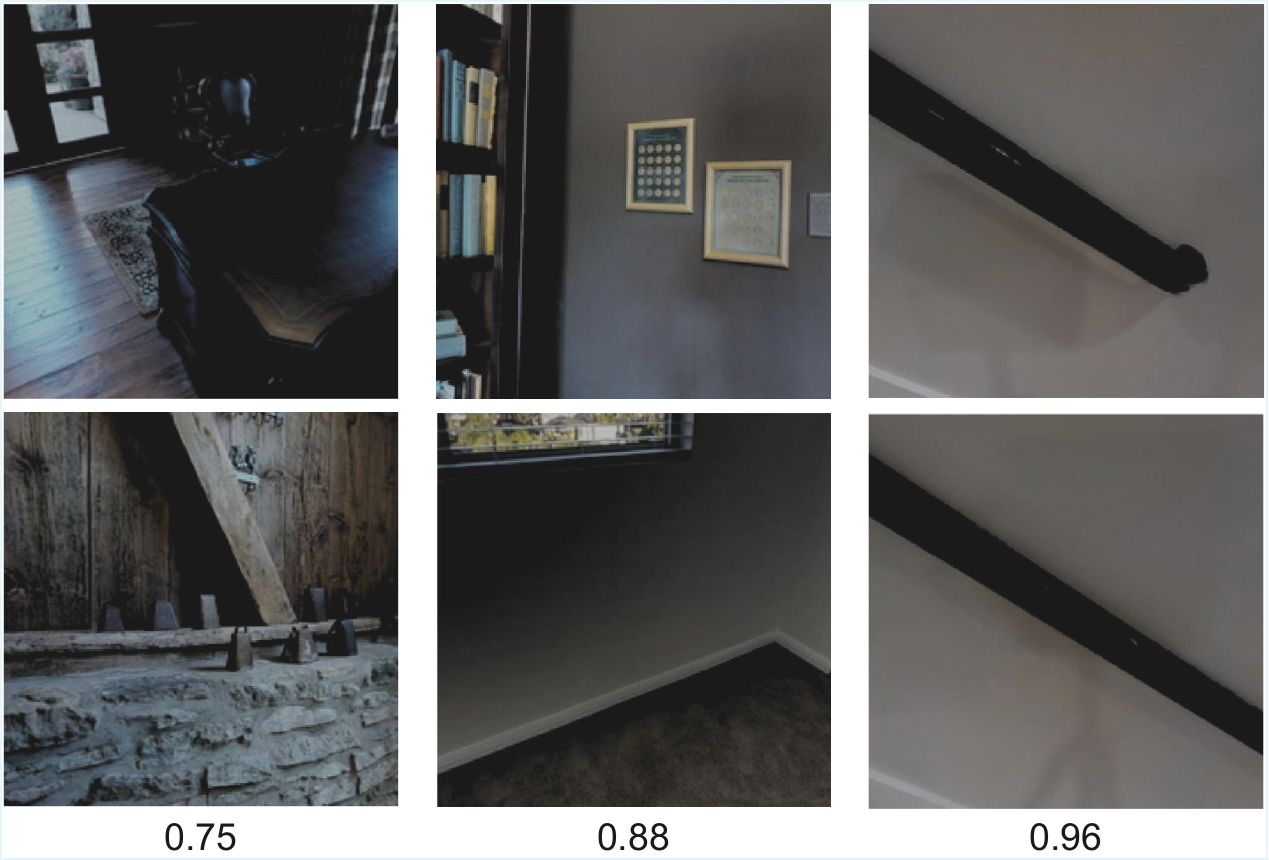}
    \vspace{-2em}
    \caption{\textbf{Cosine similarity of different views} from R2R. Comparisons are made between the upper and lower images.}
    \label{fig:cosims-comparison}
\end{figure}

\topic{LSH.}
Finally, for LSH, we analyze what types of semantic information are lost when replacing embeddings by comparing images with different cosine similarities. 
Figure~\ref{fig:cosims-comparison} shows three representative examples.
When the cosine similarity is low ($<$0.85), the views typically depict entirely different scenes or locations (e.g., the left pair of views). 
Reusing the corresponding embeddings in this case would result in a complete loss of information, substantially degrading agent performance. 
In contrast, when the cosine similarity is above 0.85---the threshold used in Sec~\ref{subsec:effectiveness}---the views are generally much more semantically similar.
For instance, the middle pair of views both show a wall of similar color, while the right pair depicts a slightly shifted angle of the same handrail.
The embeddings of such images likely encode similar information with minimal loss, which explains the limited impact of our LSH technique on performance.

\begin{table}
\centering
\includegraphics[width=1.0\linewidth]{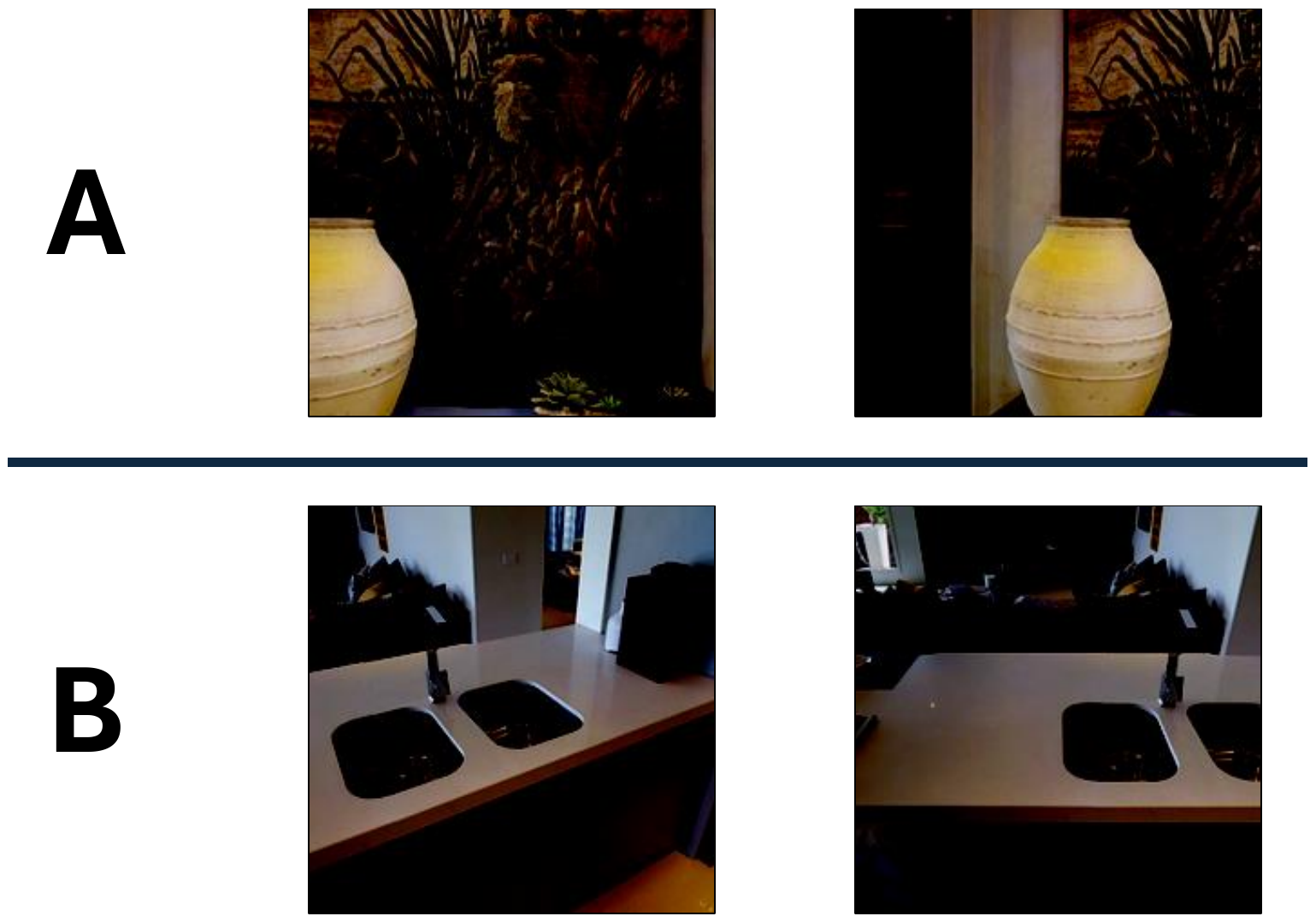}
\caption{%
    \textbf{Two sets of example views (A and B)}
    demonstrating non-identical but similar views 
    that have been slightly shifted during navigation.}
\label{fig:two-sets-of-images}
\vspace{2.0em}
\adjustbox{max width=\linewidth}{
    \begin{tabular}{@{}lcc}
    \toprule
    \textbf{Simiarlity Metrics} & \textbf{Set A} & \textbf{Set B} \\ \midrule \midrule
    \textbf{SSIM}~\cite{wang2004image} & 0.24 & 0.32 \\
    \textbf{FSIM}~\cite{zhang2011fsim} & 0.26 & 0.27 \\
    \textbf{LPIPS}~\cite{zhang2018unreasonable} & 0.55 & 0.62 \\ \midrule
    \textbf{SURF}~\cite{bay2006surf} & 0.31 & 0.32 \\
    \textbf{SIFT}~\cite{lowe2004distinctive} & 0.45 & 0.37 \\ 
    \textbf{ORB}~\cite{rublee2011orb} & 0.07 & 0.19 \\ \bottomrule
    \end{tabular}
}
\vspace{0.5em}
\caption{%
    \textbf{Similarity scores measured on Set A and B.}
    We test 6 different similarity metrics.}
\label{tbl:qualitative-comparison}
\end{table}

\section{Similarity Metrics Comparison}
\label{appendix:similarity-metrics}

Other than the three similarity metrics we use in Sec~\ref{subsec:ablation},
we test three additional metrics for comparison:
SURF~\cite{bay2006surf}, SIFT~\cite{lowe2004distinctive}, and ORB~\cite{rublee2011orb}.
These are feature detection and description algorithms 
designed to identify and match keypoints in images.
The similarity scores are computed by 
dividing the number of matching keypoints 
by the minimum number of keypoints detected in the two images.
We test all six algorithms on two sets of scenes,
reflecting shifts caused by an agent's changing perspectives during navigation.

Figure~\ref{fig:two-sets-of-images} illustrates the two scenes,
and Table~\ref{tbl:qualitative-comparison} summarizes the quantitative comparison.
Among the three metrics we employ for our main evaluation, 
LPIPS demonstrates a higher similarity measure of approximately 60\% for both sets.
In contrast, SSIM and FSIM are less effective
at capturing the similarity between views in Sets A and B.
The three additional metrics (SURF, SIFT, and ORB)
are also ineffective in providing reliable similarity scores for both image sets A and B.
Our qualitative comparison of different similarity metrics applied to sets of similar scenes highlights the challenges in accurately identifying true visual similarity.
We believe that an accurate measure of scene similarity
is crucial for further reducing the computational demands of a VLN agent,
and we leave this for future work.

\section{Performance-Efficiency Trade-off Analysis}
\label{appendix:performance_tradeoff}

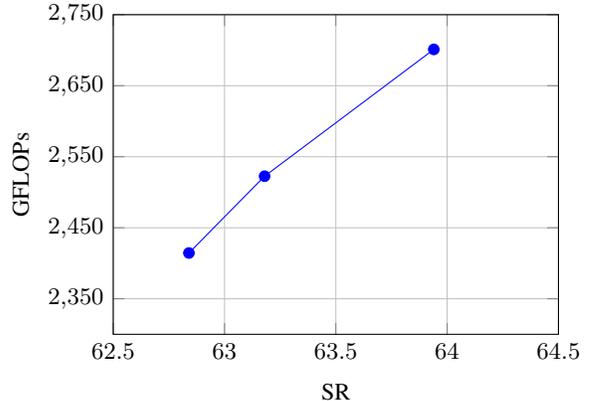
\begin{figure}[ht]
    \centering
    \adjustbox{max width=\linewidth}{
    \begin{tikzpicture}[font=\small]
        \begin{axis}[
            xlabel={SR},
            ylabel={GFLOPs},
            xmin=62.5, xmax=64.5,
            ymin=2300, ymax=2750,
            xtick={62.5, 63, 63.5, 64, 64.5},
            ytick={2350, 2450, 2550, 2650, 2750},
            grid=both,
            grid style={line width=.1pt, draw=gray!10},
            major grid style={line width=.2pt, draw=gray!50},
            legend pos=north west,
            width=0.9\linewidth,
            height=0.7\linewidth,
        ]
        \addplot[
            color=blue,
            mark=*,
            ]
            coordinates {
                (62.84, 2414.46)
                (63.18, 2522.58)
                (63.94, 2701.11)
            };
        \end{axis}
    \end{tikzpicture}
    }
    \caption{\textbf{Trade-off between Performance (SR) and GFLOPs.}}
    \label{fig:performance_vs_gflops}
\end{figure}

In order to illustrate our tunable performance-efficiency trade-off, we show that even when limiting the performance drop to under 5\%, our input adaptive inference method applied to the HAMT agent achieves significant computational savings. 
For reference, the baseline HAMT model achieves a SR of 66.16 with a computational cost of 4763.24 GFLOPs.
Figure~\ref{fig:performance_vs_gflops} shows that with a 3--5\% drop in SR, we still manage to achieve 43--50\% savings in GFLOPs.
These results were tested on the R2R validation unseen dataset.

\section{Related Work on Model Compression}
\label{appendix:model-compression}

Research has proposed an orthogonal approach 
to reduce the computational demands and memory footprint of deep-learning models:
\emph{model compression}.
Quantization and pruning are the leading practice in model compression.
Quantization ~\cite{jacob2018quantization,choi2018pact,louizos2018relaxed, bhalgat2020lsq+, uhlich2019mixed, banner2019post, choukroun2019low, li2021brecq, nagel2020up} transforms the memory representation of model parameters 
from 32-bit floating point numbers to a lower-bit integers (e.g., 4-bit integers), thereby making it more storage efficient and lowering memory usage.
Pruning
~\cite{molchanov2016pruning, fan2019reducing, fang2023depgraph, nova2023gradient, han2015learning, han2015deep, hoang2023revisiting}
aims to create sparse models by removing parameters that are less important for maintaining performance, effectively reducing model size and computation.

While quantization and pruning have been demonstrated in simpler unimodal encoder settings for image and text, they are much more challenging in vision-language model(VLM) settings~\cite{wang2022efficientvlm, sun2024p4q} and largely unexplored in VLN.
~\cite{wang2022efficientvlm}
highlighted the challenges of pruning VLMs due to the unequal weighting of visual and linguistic modalities. 
They mitigated this by using a modal-adaptive approach, adjusting pruning ratios across different model components based on downstream task sensitivity.
Similarly,~\cite{sun2024p4q}
demonstrated that naively applying post-training quantization to CLIP caused significant performance degradation, which they addressed by introducing prompt tuning and alignment modules.

We expect similar challenges to be exhibited by VLN agents, if not exacerbated.
VLN models, in addition to processing language and visual modalities, involve sequential decision-making dependent on actions taken at each time step.
We anticipate the complex interactions between these information sources to require careful consideration while adapting model compression techniques.
Future research on such techniques can be superposed along with our input-adaptive inference method to develop highly efficient models with an acceptable performance trade-off.

\section{Generalizability to Other EAI Settings}
\label{appendix:generalization}

Here, we discuss the applicability of our proposed techniques to additional embodied AI (EAI) settings.

\topic{Physical-world deployment.}
The ultimate goal of VLN research is the effective and efficient deployment of agents in the physical world.
We believe our computational efficiency generalizes to real-world deployment, as physical embodied agents typically comprise building a harness around agents trained in discrete environments~\cite{wang2024simtoreal, anderson2021simtoreal}. 
Several challenges in this process include waypoint prediction, building navigation graphs, the visual domain gap, and latency. 
We address these in our work.
We study the first two in our continuous environment experiment (Sec~\ref{subsec:continuous-results}) and the visual domain gap with natural visual degradations in Sec~\ref{subsec:robustness}. 
Our work offers a direct mechanism to address latency, which can lower barriers to practical real-world deployment.

\topic{General embodied settings.}
While our approach is designed for panoramic observations, it generalizes to other EAI settings. 
Panoramas are extensively used in non-VLN tasks, e.g., visual navigation~\cite{wasserman2023last}, humanoid robots~\cite{zhang2025humanoidpano}, and autonomous driving~\cite{zhu2021vigor}. 
We expect high transferability to any setting employing panoramas. 
Generally, panoramic observations provide a wider scene context that can be valuable for decision making, albeit at the cost of computations. 
Our method alleviates this limitation and can facilitate wider use of panoramas for embodied AI.

\end{document}